\documentclass{article}

\PassOptionsToPackage{square, numbers}{natbib}
    \usepackage[final]{neurips_2022}

\RequirePackage{algorithm}
\RequirePackage{algorithmic}
\usepackage[utf8]{inputenc} % allow utf-8 input
\usepackage[T1]{fontenc}    % use 8-bit T1 fonts
\usepackage{hyperref}       % hyperlinks
\usepackage{url}            % simple URL typesetting
\usepackage{booktabs}       % professional-quality tables
\usepackage{amsfonts}       % blackboard math symbols
\usepackage{nicefrac}       % compact symbols for 1/2, etc.
\usepackage{microtype}      % microtypography
\usepackage[table, dvipsnames]{xcolor}
\usepackage[normalem]{ulem}
\usepackage{caption}
\usepackage{subcaption}
\usepackage{cite}
\usepackage{pifont}
\usepackage{dsfont}
\usepackage{bbm}
\usepackage{xspace}
\usepackage{mathtools}
\usepackage{amsthm}
\usepackage{enumitem}
\usepackage{array,multirow}
\usepackage[capitalize,noabbrev]{cleveref}

\usepackage{wrapfig}

\usepackage{enumitem} % Required for list customisation
\setlist{noitemsep} % No spacing between list items

\usepackage{xpunctuate}
\providecommand{\eg}[0]{\xperiodafter{e.g}}

\providecommand{\ie}[0]{\xperiodafter{i.e}}

\providecommand{\etal}[0]{\xperiodafter{\emph{et al}}}
\providecommand{\vs}[0]{\xperiodafter{\emph{vs}}}
\providecommand{\etc}[0]{\xperiodafter{etc}}

\DeclareMathAlphabet{\nicecal}{OMS}{zplm}{m}{n}

\DeclareMathOperator*{\argmin}{arg\,min}

\theoremstyle{plain}

\theoremstyle{definition}

\theoremstyle{remark}

\renewcommand{\cite}[1]{\citep{#1}}

\def\Vec#1{{\boldsymbol{#1}}}

\newcommand{\ignore}[1]{}
\definecolor{Gray}{gray}{0.9}

\title{Rethinking Generalization in Few-Shot Classification
}

\author{%
    Markus Hiller\thanks{Joint first authorship \ignore{$\;$ \textdagger~Corresponding author}}\textsuperscript{$\;\;$\ignore{\textdagger}1} \quad Rongkai Ma\textsuperscript{$*$2} \quad Mehrtash Harandi\textsuperscript{2} \quad Tom Drummond\textsuperscript{1}\\
    \vspace{-2.5mm}\\
    \textsuperscript{1}School of Computing and Information Systems, The University of Melbourne\\
    \textsuperscript{2}Department of Electrical and Computer Systems Engineering,  Monash University\\
    \texttt{markus.hiller@student.unimelb.edu.au} \\
    \texttt{\{rongkai.ma, mehrtash.harandi\}@monash.edu}\\
    \texttt{tom.drummond@unimelb.edu.au}
}

\begin{document}
\setcitestyle{square}

\maketitle

\begin{abstract}

Single image-level annotations only correctly describe an often small subset of an image's content, particularly when complex real-world scenes are depicted. While this might be acceptable in many classification scenarios, it poses a significant challenge for applications where the set of classes differs significantly between training and test time.
In this paper, we take a closer look at the implications in the context of \textit{few-shot learning}.
Splitting the input samples into patches and encoding these via the help of Vision Transformers allows us to establish semantic correspondences between local regions across images and independent of their respective class. 
The most informative patch embeddings for the task at hand are then determined as a function of the support set via online optimization at inference time, additionally providing visual interpretability of `\textit{what matters most}' in the image.
We build on recent advances in unsupervised training of networks via masked image modelling to overcome the lack of fine-grained labels and learn the more general statistical structure of the data while avoiding negative image-level annotation influence, \textit{aka} supervision collapse.
Experimental results show the competitiveness of our approach, achieving new state-of-the-art results on four popular few-shot classification benchmarks for $5$-shot and $1$-shot scenarios. 
%The code is publicly available at \url{https://github.com/mrkshllr/FewTURE}.
\end{abstract}
\section{Introduction}

\begin{figure*}[t] 
    \begin{center}
        \includegraphics[width=1\linewidth]{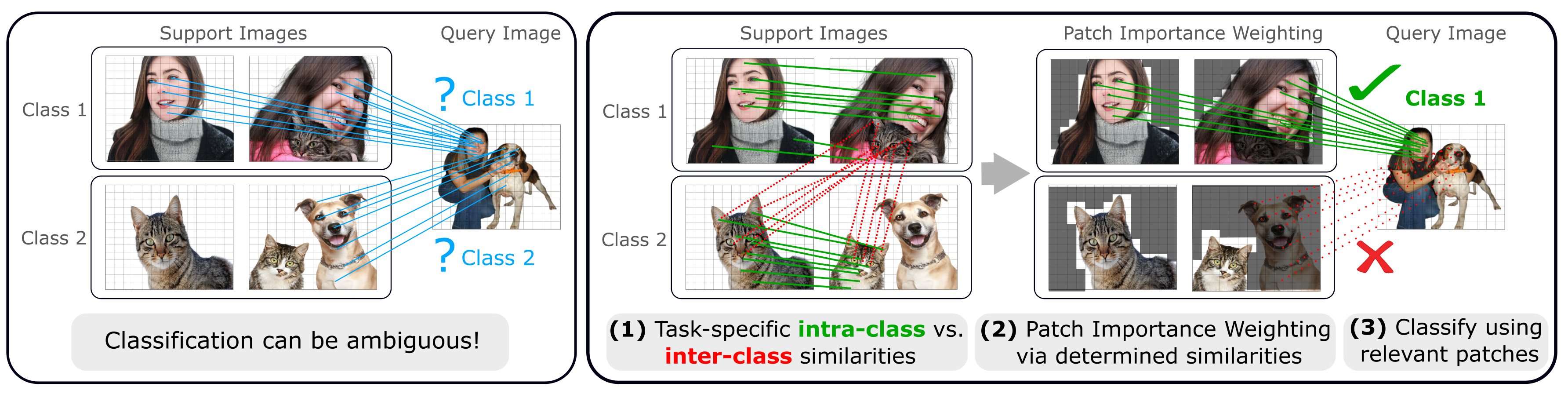}
    \end{center}
    \vskip -0.08in
    \caption{\textbf{Tackling classification ambiguity \ignore{of single image-level annotations }by interpreting images in context.} (Left): Labels assigned to real-world images with multiple entities only correctly describe a subset of the depicted content, leading to ambiguous classification results. (Right): Leveraging intra- and inter-class similarities and differences across the support set allows our method to determine the importance of each individual patch at inference time, \ie, to find out '\textit{what matters most}' in each image. This information is then used to reweight support-query similarities and resolve ambiguity.
    }
    \label{fig:motivation}
    \vskip -0.12in
\end{figure*}

Images depicting real-world scenes are usually comprised of several different entities, \eg, a family walking their dog in a park surrounded by trees, or a person patting their dog (\cref{fig:motivation}). Nevertheless, popular computer vision datasets like ImageNet~\cite{russakovsky_2015imagenet} assign a single \textit{image-level} annotation to classify their entire content. Hence, such a label only correctly applies to an often small subset of the actual image.
As a result, models trained on such data via gradient-based methods learn to ignore all seemingly irrelevant information, particularly entities that occur across differently labelled images.
While this might be acceptable for conventional classification methods that encounter a diverse number of training examples for all classes they are expected to distinguish, it poses a major but often overlooked challenge for applications where the set of classes differs between training and test time. One such affected area is few-shot learning (FSL) where approaches are expected to correctly classify entirely new classes at test time that have never been encountered during training, just by being provided with a few (\eg, one or five) samples for each of these new categories. 
During test time, entities that have not been part of the set of training classes and have possibly been perceived as irrelevant might very well be part of the set of test classes -- yet, the method was taught to ignore these. Similarly, a method might overemphasize the importance of certain image patterns learned during training that are however of no relevance for the test classes, resulting in \ignore{classification failure aka }\textit{supervision collapse}~\cite{doersch2020_crosstransformers}. 

Few previous works~\cite{doersch2020_crosstransformers, hou2019_crossattention} partially tackle the above challenges. CTX~\cite{doersch2020_crosstransformers} proposes to learn the spatial and semantic alignment between CNN-extracted query and support features using a Transformer-style attention mechanism. The authors further show that self-supervised learning tasks (\ie, SimCLR) can be integrated into episodic training along with normal supervised tasks to learn more generalized features, which benefits solving unseen tasks and mitigates supervision collapse. CAN~\cite{hou2019_crossattention} achieves this in a similar manner by performing cross-attention between class prototypes and query feature maps, highlighting the region of a feature map important for classification during inference. While both methods propose important contributions towards tackling supervision collapse, there exist important drawbacks. Firstly, both methods build their ideas around aligning prototypes based on each query. Such prototypes are merely class-aware and ignore all inter-class information present in the support set -- a part that has however been shown to be crucial for few-shot learning~\cite{oreshkin2018_tadam_fc100,vinyals2016_miniimagenet}. Furthermore, learning query-aligned class representations requires performing the same operation \ignore{multiple (equal to the number of query samples) times}for each query, rendering such approaches rather inefficient at inference time.

Summarizing our and previous works' observations, we aim to address what we see as the two main criteria: 1) building an understanding about an image's structure and content that \textit{generalizes} towards new classes, and 2) providing the ability to interpret the provided samples \textit{in context}, \ie, finding the intra-class similarities and inter-class differences while jointly considering all available information.  

\textbf{Our work.}\footnote{Our code is publicly available at \url{https://github.com/mrkshllr/FewTURE}} 
To alleviate the negative influence of image-level annotations and to avoid supervision collapse, we decompose the images into patches representing local regions, each having a higher likelihood of being dominated by only one entity. To overcome the lack of such fine-grained annotations we employ self-supervised training with Masked Image Modelling as pretext task~\cite{zhou2021_ibot} and use a Vision Transformer architecture~\cite{dosovitskiy2020_vit} as encoder due to its patch-based nature. We build our classification around the concept of learning task-specific similarity between local regions as a function of the support set at inference time. To this extent, we first create a prior similarity map by establishing semantic patch correspondences between all support set samples irrespective of their class, \ie, also between entities that might not be relevant or potentially even harmful for correct classification (\cref{fig:motivation},~step~(1)). Consider the depicted support set with only two classes: `person' and `cat'. The lower-right image is part of our support set for `cat' -- and the dog just happens to be in the image. Now in the query sample that shall be classified, the image depicts a person patting their dog. We will thus correctly detect a correspondence of the two dogs across those two images, as well as between the person patches and the other samples of the person support set class. While the correspondences between the person regions are helpful, there is no `dog' class in the actual support set (\ie, `dog' is out-of-task information), rendering this correspondence harmful for classification since it would indicate that the query is connected to the image with the `cat' label.
This is where our \textit{token importance weighting} comes into play. We infer an importance weight for each token based on its contribution towards correct classification of the other support set samples, actively strengthening intra-class similarities and inter-class differences by jointly considering all available information -- in other words, we learn which tokens `help' or `harm' our classification objective (\cref{fig:motivation},~step~(2)).
These importance-reweighted support set embeddings are then used as basis for our similarity-based query sample classification~(step~(3)). 
Our main contributions include the following:
\begin{enumerate} [topsep=0pt]
    \item We demonstrate that Transformer-only architectures in conjunction with self-supervised pretraining can be successfully used in few-shot settings without the need of convolutional backbones or any additional data.
    \item We show that meta fine-tuning of Vision Transformers combined with our inner loop token importance reweighting can successfully use the supervision signal of provided support set labels while avoiding supervision collapse.
    \item We provide insights into how establishing general similarities across images independent of classes followed by our optimization-based selection at inference time can boost generalization while allowing visual interpretability at the same time, and show the efficacy of our method by achieving new state-of-the-art results on four popular public benchmarks.
\end{enumerate}

\section{Few-shot classification via reweighted embedding similarity}
\begin{figure*}[t] 
\begin{center}
\includegraphics[width=1\linewidth]{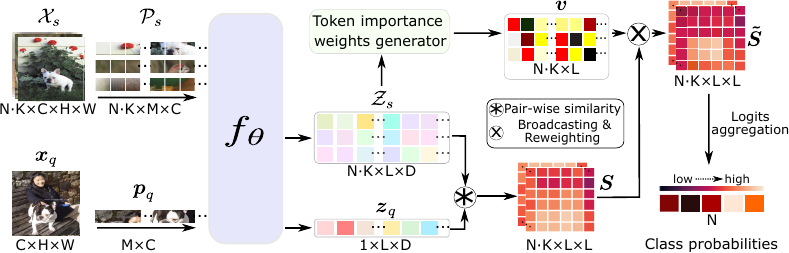}
\end{center}
\caption{\textbf{Illustration of the proposed method FewTURE}. Support and query set images are split into patches and encoded by our Transformer backbone. Classification of query set images is performed by using the reweighted similarity of the encoded patches w.r.t. the support set tokens.} 
\label{fig:framework}
\end{figure*}

We start this section by briefly introducing the problem setting we are tackling in this work: inductive few-shot classification. We then provide an overview of our proposed method \textit{FewTURE} -- \underline{Few}-shot classification with \underline{T}ransformers \underline{U}sing \underline{R}eweighted \underline{E}mbedding similarity (\cref{fig:framework}) before elaborating on the main elements in more detail.
\par
\textbf{Problem definition}.
Inductive $N$-way $K$-shot few-shot classification aims to generalize knowledge learned during training on $\mathcal{D}_{train}$ to unseen test data $\mathcal{D}_{test}$, with classes $\mathcal{C}_{train}\cap\mathcal{C}_{test}=\emptyset$, using only a few labelled samples. We follow the meta-learning protocol of previous works~\cite{vinyals2016_miniimagenet} to formulate the few-shot classification problem with episodic training and testing. An episode $\mathcal{E}$ is composed of a support set $\mathcal{X}_s = \{(\Vec{x}_s^{nk}, \boldsymbol{y}_s^{nk})|n=1,\ldots,N; k=1,\ldots,K; \boldsymbol{y}_s^{nk}\in\mathcal{C}_{train}\}$, where $\Vec{x}_s^{nk}$ denotes the $k$-th sample of class~$n$ with label $\boldsymbol{y}_s^{nk}$, and a query set $\mathcal{X}_q=\{(\Vec{x}_q^{n},\boldsymbol{y}_q^{n})|n=1,\ldots,N\}$, where $\Vec{x}_q^{n}$ denotes a query sample\footnote{Without loss of generality, we present our method for the case of one query sample per class to improve ease of understanding. The exact number of query samples per class is generally unknown in practice.} of class~$n$ with label $\boldsymbol{y}_q^n$. 

\subsection{Overview of FewTURE}% \textcolor{red}{\textit{or} Looking into (the) FewTURE? (haha)}}
As depicted in \cref{fig:framework}, we encode the image patches $\mathcal{P}_{s}$ of the support set samples along with the query sample patches~$\mathbf{p}_q$ via~$f_{\theta}$ and obtain corresponding sets of tokens $\mathcal{Z}_s$ and $\boldsymbol{z}_q$, respectively\footnote{Note that some tensor shapes in the illustrations might differ from the equations for ease of visualization.}.  
It is to be noted that while we choose to illustrate our method via the use of one single query sample, the classification of all query samples is computed at the same time in one single pass in practice. 
We retrieve our `prior' correspondence map~$\boldsymbol{S}$ expressing the token-wise similarity between the encoded semantic content of the local regions in the query sample and all patches of all support samples, allowing us to consider all available information jointly without incurring information loss due to averaging or similar operations. 
This \textit{`prior' similarity map} represents correspondences between regions of samples irrespective of their individual class, \ie also between entities that might not be relevant or potentially even harmful for correct classification.
Using the annotated support set samples, we now infer a task-specific importance weight factor~$v^j$ for each support token~$z_{s}^{j}$ representing its contribution to correctly classify other samples in the support set via online optimization at inference time (\cref{sec:token_importance_opt}). We then reweight the prior similarities to obtain our classification result for the query sample, jointly considering all available information.

\subsection{Self-supervised pretraining against supervision collapse}
To overcome the problem of supervision collapse induced by image-level annotations, we split the input images into smaller parts where each region has a higher likelihood of only containing one major entity and hence a more distinct semantic meaning. Since no labels are available for this more fine-grained data, we encode the information of each local region via an unsupervised method.

We build our approach around the recently introduced idea of using \textit{Masked Image Modeling} (MIM)~\cite{bao2021_beit,zhou2021_ibot} as a pretext task for self-supervised training of Vision Transformers. 
In contrast to previous unsupervised approaches~\cite{caron2021_dino, chen2021_mocov3} which focused mainly on global image-level representations, MIM randomly masks a number of patch embeddings (\textit{tokens}) and aims to reconstruct them given the remaining information of the image. 
The \ignore{in this way }introduced token constraints help our Transformer backbone to learn an embedding space that yields semantically meaningful representations for each individual image patch.
We then leverage the information of the provided labels through fine-tuning the pretrained backbone in conjunction with our inner loop token importance weighting described in the following sections while successfully avoiding supervision collapse (see experimental results in~\cref{sec:exp_selfsup_generalization}).

\subsection{Classification through reweighted token similarity}
As illustrated in \cref{fig:framework}, we split each input image $\boldsymbol{x}\in \mathbb{R}^{H\times W \times C}$ into a sequence of $M\!=\nicefrac{H\cdot W}{P^2}$ patches $\mathbf{p}=\{p^i\}^{M}_{i=1}$, with each patch $p^i \in \mathbb{R}^{P^2\times C}$. We then flatten and pass all patches of the support and query images as input to the Transformer architecture, obtaining the set of support tokens~$\mathcal{Z}_s\!=\!f_{\theta}(\mathcal{P}_s)$ with~$\mathcal{Z}_{s}\!=\!\{\Vec{z}_s^{nk}|n\!=\!1,\ldots,N,k\!=\!1,\ldots,K\}, \Vec{z}_{s}^{nk}\!=\!\{z_s^{nkl}|l\!=\!1,\ldots,L;z_{s}^{nkl}\in\mathbb{R}^{D}\}$ and query tokens~$\Vec{z}_q\!=\!f_{\theta}(\mathbf{p}_q)$ with $\Vec{z}_q\!=\!\{z_q^l|l\!=\!1,\ldots,L;z_q^{l}\in\mathbb{R}^{D}\}$.
Vision Transformers like ViT~\cite{dosovitskiy2020_vit,touvron2021_deit} satisfy~$L=M$ whereas hierarchical Transformers like Swin~\cite{liu2021_swin} generally emit a reduced number of tokens $L<M$ due to internal merging strategies. 

Having obtained all patch embeddings, we establish semantic correspondences by computing the pair-wise patch similarity matrix between the set of support tokens~$\mathcal{Z}_{s}$ and query tokens~$\boldsymbol{z}_q$ as
$\boldsymbol{S}\in\mathbb{R}^{N\cdot K\cdot L\times L}$, where each element in $\boldsymbol{S}$ is obtained by $s_{nk}^{l_s,l_q}={\mathrm{sim}}(z_s^{nkl_s},z_q^{l_q})$, where $l_s\!=\!1,\ldots,L$ and $l_q\!=\!1,\ldots,L$. Note that local image regions representing similar entities exhibit higher scores. While any distance metric can be used to compute the similarity ($\mathrm{sim}$), we found cosine similarity to work particularly well for this task.
We then use the task-specific token importance weights~$\boldsymbol{v} \in \mathbb{R}^{N\cdot K \cdot L \times 1}$ inferred via online optimization based on the annotated support set samples (see \cref{sec:token_importance_opt}) to reweight the similarities through column-wise addition and obtain our task-specific similarity matrix as ${\boldsymbol{\Tilde{S}}}= \boldsymbol{S} + \left[\boldsymbol{v} \cdot \mathbbm{1}^{1\times L}\right]$, with elements~$\Tilde{s}_{nk}^{l_s,l_q}$. Note that this addition of our reweighting logits corresponds to multiplicative reweighting in probability space. We temperature-scale the adapted similarity logits with $\nicefrac{1}{\tau_S}$ and aggregate the token similarity values across all elements belonging to the same support set class via a \textit{LogSumExp} operation, \ie aggregating
$K\cdot L^2$ logits per class followed by a softmax -- resulting in the final class prediction~$\hat{\boldsymbol{y}}_q$ \ignore{\in \mathbb{R}^N }for the query sample~$\boldsymbol{x}_q$ as
\begin{equation} \label{eq:pred}
    \hat{\boldsymbol{y}}_q = \mathrm{softmax}\left(\left\{\hat{y}_q^n\right\}_{n=1}^N\right)= \mathrm{softmax}\left(\left\{\log\sum_{k=1}^{K}\sum_{l_q=1}^{L}\sum_{l_s=1}^{L}\mathrm{exp}\left(\nicefrac{\Tilde{s}_{nk}^{l_s,l_q}}{\tau_S}\right)\right\}_{n=1}^N\right)\,.
\end{equation}
\subsection{Learning token importance at inference time}
\label{sec:token_importance_opt}
\begin{figure*}[t] 
\begin{center}
\includegraphics[width=1\linewidth]{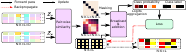}
\end{center}
\caption{\textbf{Inner loop token importance weight generator}. The most helpful tokens for the task at hand are determined as a function of the support set via inner-loop optimization at inference time by reweighting all token similarities based on their contribution towards a correct classification result. } 
\label{fig:tiwg}
\vspace{-0.7em}
\end{figure*}

We use all samples of a task's support set together with their annotations to learn the importance for each individual patch token via online optimization at inference time. 
As depicted in \cref{fig:tiwg}, we formulate the same classification objective as in the previous section but aim to classify the support set samples instead of a query sample. In other words, we use two versions of the support set samples: one with labels as `support'~$\mathcal{Z}_s$ and one without as `pseudo-query'~$\mathcal{Z}_{sq}$, and obtain the similarity matrix  $\boldsymbol{S}_s\in\mathbb{R}^{N\cdot K\cdot L\times N\cdot K\cdot L}$. The token importance weights are initialized to~$\boldsymbol{v}^0=\boldsymbol{0} \in \mathbb{R}^{N\cdot K\cdot L \times 1} $ and column-wise added to form $\Tilde{\boldsymbol{S}}_s= \boldsymbol{S}_s + \left[\boldsymbol{v}^0 \cdot \mathbbm{1}^{1\times N \cdot K \cdot L}\right]$. \par
The goal is now to determine which tokens of~$\mathcal{Z}_s$ are most helpful in contributing towards correctly classifying~$\mathcal{Z}_{sq}$, and which ones negatively affect this objective. To prevent tokens from simply classifying themselves, we devise the following strategy for our $N$-way $K$-shot tasks.
For scenarios with $K>1$ samples per class, we apply block-diagonal masking with blocks of size $L\times L$ to the similarity matrix~$\Tilde{\boldsymbol{S}}_s$ -- meaning that we enforce classification of each token in~$\mathcal{Z}_{sq}$ exclusively based on information from other images. Since there are no other in-class examples available in $1$-shot scenarios, we slightly weaken the constraint and apply local masking in an $m\times m$ window around each patch, forcing the token to be classified based on the remaining information in the image. We found a local window of $m=5$ to work well throughout our experiments for both architectures.

We use temperature-scaling and aggregate all modified similarity logits across all elements belonging to the same support set class of the annotated~$\mathcal{Z}_s$ for each element~$\boldsymbol{z}_{sq}^{nk}$, apply a softmax operation and obtain the predicted class probabilities~$\hat{\boldsymbol{y}}^{nk}_s$ for each support set sample (\textit{cf.} \cref{eq:pred}). Given that the prediction is now dependent on the initialized token similarity weights~$\boldsymbol{v}$, we can formulate an online optimization objective by using the support set labels~$\boldsymbol{y}^{nk}_s$ as
\begin{equation}
    \argmin_{\boldsymbol{v}} \sum_{n=1}^{N}\sum_{k=1}^{K}\mathcal{L}_{\mathrm{CE}}\left(\boldsymbol{y}^{nk}_s,\;\hat{\boldsymbol{y}}^{nk}_s\left(\boldsymbol{v}\right)\right) \,.
\end{equation}
It is to be noted that by using column-wise addition of~$\boldsymbol{v}$, we share the importance weights of each support token across all `pseudo-query' tokens and thus constrain the optimization to jointly learn token importance with respect to all available information. In other words, task-specific intra-class similarities will be strengthened while inter-class ones will be penalized accordingly. We further do not require any second order derivatives like other methods during meta fine-tuning, since the optimization of~$\boldsymbol{v}$ is decoupled from the network's parameters. 

\section{Experiments and discussion}
\subsection{Implementation details} \label{sec:impl_details}
Our training strategy is divided into two parts: self-supervised pretraining followed by meta fine-tuning. It is to be noted that each architecture is exclusively trained on the training set data of the respective dataset that is to be evaluated, and no additional data is used. \par
\textbf{Datasets}.
We train and evaluate our methods using four popular few-shot classification benchmarks, namely \textit{mini}ImageNet~\cite{vinyals2016_miniimagenet}, \textit{tiered}ImageNet~\cite{ren2018_tieredimagenet}, CIFAR-FS~\cite{bertinetto2019_cifarfsl} and FC-100~\cite{oreshkin2018_tadam_fc100}.\par
\textbf{Architectures}. 
We compare two different Transformer architectures in this work, the monolithic ViT architecture~\cite{dosovitskiy2020_vit,touvron2021_deit} in its `small' form (\textit{ViT-small} \textit{aka DeiT-small}) and the multi-scale Swin architecture~\cite{liu2021_swin} in its `tiny' version (\textit{Swin-tiny}).\par
\textbf{Self-supervised pretraining}. 
We employ the strategy proposed by~\cite{zhou2021_ibot} to pretrain our Transformer backbones and mostly stick to the hyperparameter settings reported in their work. Our ViT and Swin architectures are trained with a batch size of $512$ for $1600$ and $800$ epochs, respectively. We use 4 Nvidia A100 GPUs with 40GB each for our ViT and 8 such GPUs for our Swin models. \par
\textbf{Meta fine-tuning}.
We use meta fine-tuning to further refine our embedding space by using the available image-level labels in conjunction with our token-reweighting method. We generally train for up to $200$~epochs but find most architectures to converge earlier. We evaluate at each epoch on 600 randomly sampled episodes from the respective validation set to select the best set of parameters. During test time, we randomly sample 600 episodes from the test set to evaluate our model. \par
\textbf{Token importance reweighting and classifier}. We use cosine similarity to compute~$\boldsymbol{S}$. While the temperature~$\tau_S$ used to scale the logits can be learnt during meta fine-tuning, we found $\tau_S = \nicefrac{1}{\sqrt{d}}$ to be a good default value (or starting point if learnt, see supplementary material). We use SGD as optimizer with a learning rate of 0.1 for the token importance weight generation.\par
Please refer to supplementary material of this paper for a more detailed discussion regarding datasets, implementation and hyperparameters.

\subsection{Self-supervised pretraining and token-reweighted fine-tuning improve generalization}
\label{sec:exp_selfsup_generalization}
\begin{figure}[t] 
    \begin{center}
        \includegraphics[width=1\linewidth]{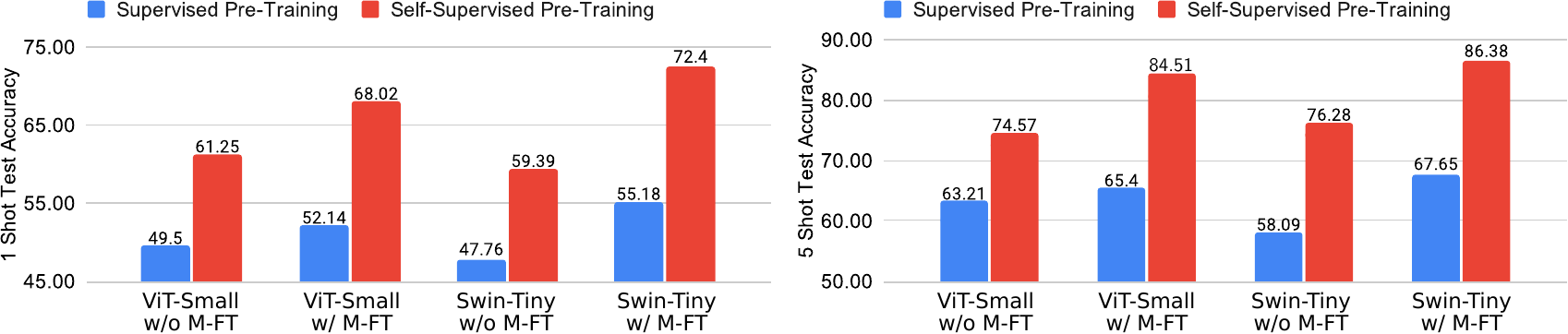}
    \end{center}
    \caption{\textbf{Supervised vs. self-supervised pretraining.} Average test accuracies on \textit{mini}ImageNet for our method with different pretraining methods, with (\textit{w/}) and without (\textit{w/o}) meta fine-tuning (M-FT).}  
    \label{fig:selfsup_sup}
\end{figure}

\begin{figure}[t] 
\vspace{0.5em}
    \begin{center}
        \includegraphics[width=1\linewidth]{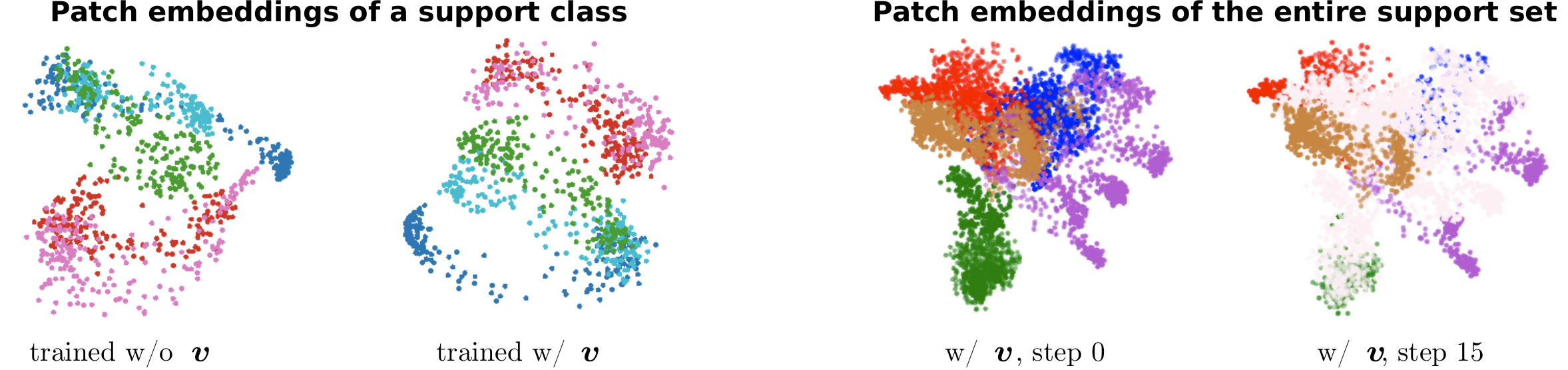}  
    \end{center}
    \caption{\textbf{Instance and class embeddings.} Visualised are the projected tokens of 5 instances of the same novel support set class (left) and of the entire support set (right). From left to right: Instance embeddings meta-trained using our classifier without task-specific token reweighting (`\textit{w/o} $\boldsymbol{v}$') \vs trained with 15 reweighting steps (`\textit{w/} $\boldsymbol{v}$'); Embeddings of entire 5-way 5-shot support set obtained by our approach trained with 15 steps, displayed at reweighting step 0 \vs. step 15. (PCA projection)}
    \label{fig:TSNE}
    \vspace{-0.5em}
\end{figure}

In this section, we investigate the influence of self-supervised pretraining compared to its supervised counterpart using the provided image-level labels (de-facto standard in most current state-of-the-art methods). We only vary the training strategy of our backbone and use our introduced token similarity-based classifier with 15 inner-loop reweighting steps for all experiments. \cref{fig:selfsup_sup} illustrates that both ViT-small and Swin-tiny with self-supervised pretraining alone (\textit{w/o} meta fine-tuning) learn more generalizable features than their respective supervised versions for both $1$-shot (left-hand side) and $5$-shot (right-hand side) scenarios. In fact, across all cases \textit{w/o} meta fine-tuning, the self-supervised pretrained Transformers outperform their supervised counterparts by more than $10\%$ and up to a significant $18.19\%$ in the case of Swin-tiny. This clearly demonstrates that our self-supervised pretraining strategy explores information that is richer and beyond the labels. 
Another interesting insight is that the meta fine-tuning stage does not improve supervised backbones as much as their unsupervised counterparts. Specifically after unsupervised pretraining, meta fine-tuning of the Transformers in conjunction with our token-reweighting strategy is able to boost the performance by $6.77\%$ (ViT-small) and $13.01\%$ (Swin-tiny) for $1$-shot, and $9.94\%$ (ViT-small) and $10.10\%$ (Swin-tiny) for $5$-shot. While such significant improvements cannot be observed across the supervised networks, the Swin versions seem to generally start off lower after pretraining but benefit more from the fine-tuning than ViT. The observed results clearly indicate that our token-reweighted fine-tuning strategy is able to further improve the generalization of self-supervised Transformers, thus performing better on the novel tasks of the unseen test set. 
Figure~\ref{fig:TSNE} additionally depicts projected views of the tokens of 5 instances from a novel class as well as the entire novel support set in embedding space. Representations obtained with our classifier seem to retain the instance information (`\textit{w/o} $\boldsymbol{v}$') and separation is improved when using token importance reweighting (`\textit{w/} $\boldsymbol{v}$'). While the projected tokens of the entire support set show partial overlap between classes as is expected due to commonalities like \eg similar background, our reweighting clearly determines the class-characteristic tokens (displayed in their original class-respective color). These results indicate that our similarity-based classifier coupled with task-specific token reweighting is able to better disentangle the embeddings of different instances from the same class as well as other classes, which prevents the network from supervision collapse and achieves the higher performance observed on the benchmarks. They further show that self-supervised pretraining is helpful but not sufficient to achieve well-separated representations without supervision collapse that are suitable for few-shot classification.

\subsection{Selecting helpful patches at inference time}
\begin{figure}[t] 
    \begin{center}
        \includegraphics[width=1\linewidth]{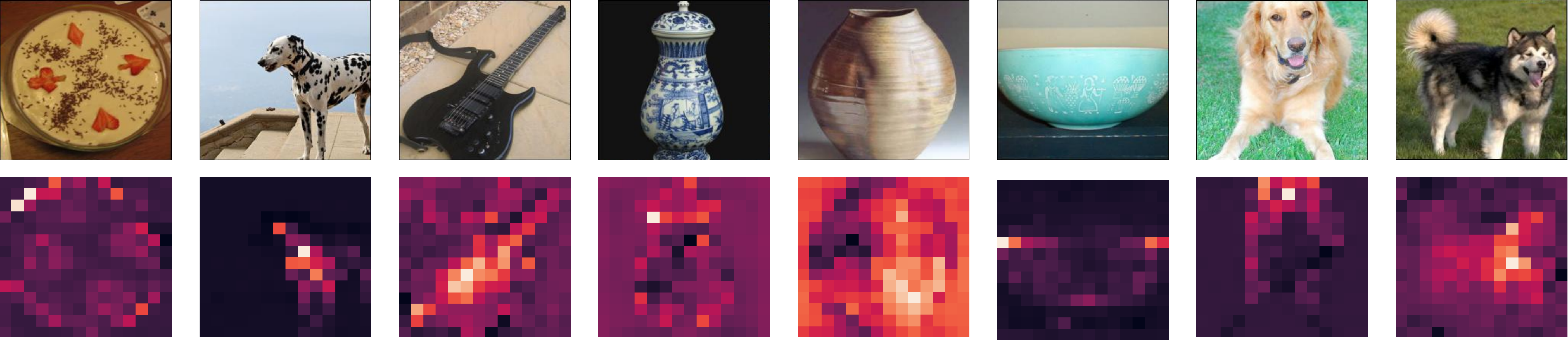}  
    \end{center}
    \caption{\textbf{Learning token importance at inference time.} Visualized importance weights learnt via online optimization for support set samples in a $5$-way $5$-shot task on the \textit{mini}ImageNet test set.} 
    \label{fig:patch_importance_v}
    \vspace{-0.7em}
\end{figure}

\cref{fig:patch_importance_v} shows a visualization of the patch importance weights~$\boldsymbol{v}$ that are learned at inference time during the inner loop adaptation for the support set images. Brighter regions represent a higher importance weight, meaning that these patches will contribute most to the classification of query samples if matches with high similarity can be found. 
Judging from the visualized weights, \textit{FewTURE} seems to consistently select characteristic regions of the depicted objects, \eg, the rim of the bowls, strings of the guitar or the dogs' facial area, and to exclude unimportant or out-of-task information. 

\begin{figure}[b]
\centering
    \begin{subfigure}[b]{0.48\textwidth}     
        \centering
         \includegraphics[width=\textwidth]{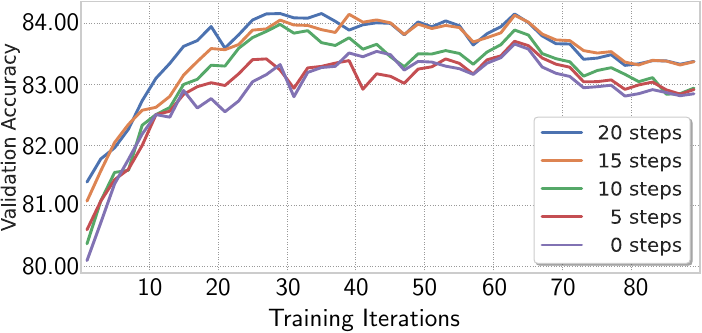}
         \caption{}
         \label{subfig:ablation_upt_steps_valacc}
     \end{subfigure}
     \hfill
     \begin{subfigure}[b]{0.48\textwidth}
         \centering
         \includegraphics[width=\textwidth]{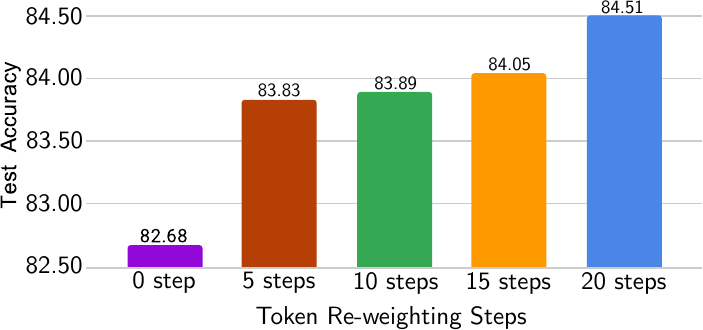}
         \caption{}
         \label{subfig:ablation_upt_steps_testacc}
     \end{subfigure}
     \caption{\textbf{Inner loop token reweighting.} Average classification accuracies on the \emph{mini}ImageNet validation set (\subref{subfig:ablation_upt_steps_valacc}), and test set (\subref{subfig:ablation_upt_steps_testacc}) for varying inner loop optimization steps, evaluated with a ViT-small backbone and SGD with 0.1 as learning rate.}
     \label{fig:ablation_upt_steps_both}
    %  \vspace{-1.5em}
\end{figure}

We further investigate the influence of the number of optimization steps in our inner loop token importance weighting method using ViT-small on \textit{mini}ImageNet.
The results in \cref{fig:ablation_upt_steps_both} indicate that increasing the steps up to 20 aligns with increased performance, both during validation and testing. While the initial increase in test accuracy when using our token reweighting (steps $>0$) is rather significant with $1.15\%$, the contribution of higher step numbers comes at the cost of higher computational complexity, and we generally found anything between 5 and 15 steps to be a good performance \vs inference-time trade-off (see supplementary material for further details).

\begin{table}[tb]
    % \vspace{-0.7em}
    \centering
    \caption{Average classification accuracy for $5$-way $1$-shot and $5$-way $5$-shot scenarios. Reported are the mean and 95\% confidence interval on the unseen test sets of \textit{mini}ImageNet~\cite{vinyals2016_miniimagenet} and \textit{tiered}ImageNet~\cite{ren2018_tieredimagenet}, using the established evaluation protocols.}
    \scalebox{0.85}
    {
    \begin{tabular}{l|c c|c c|c c}
    \specialrule{.2em}{.1em}{.1em}
    \multirow{2}{*}{\textbf{Model}}
    &\multirow{2}{*}{\textbf{Backbone}}
    &\multirow{2}{*}{\textbf{$\approx$~\# Params}}
    &\multicolumn{2}{c|}{\textbf{\emph{mini}ImageNet}}
    &\multicolumn{2}{c}{\textbf{\emph{tiered}ImageNet}}
    \\
    &
    &
    &\textbf{1-shot}
    &\textbf{5-shot}
    &\textbf{1-shot}
    &\textbf{5-shot}
    \\
    \toprule\toprule
    ProtoNet~\cite{snell2017prototypical}
    &ResNet-12
    &12.4 M
    &$62.29{\scriptstyle \pm 0.33}$
    &$79.46{\scriptstyle \pm 0.48}$
    &$68.25{\scriptstyle \pm 0.23}$
    &$84.01{\scriptstyle \pm 0.56}$
    \\
    FEAT~\cite{ye2020fewshot}
    &ResNet-12
    &12.4 M
    &$66.78{\scriptstyle \pm 0.20}$
    &$82.05{\scriptstyle \pm 0.14}$
    &$70.80{\scriptstyle \pm 0.23}$
    &$84.79{\scriptstyle \pm 0.16}$
    \\
    DeepEMD~\cite{Zhang_2020_CVPR}
    &ResNet-12
    &12.4 M
    &$65.91{\scriptstyle \pm 0.82}$
    &$82.41{\scriptstyle \pm 0.56}$
    &$71.16{\scriptstyle \pm 0.87}$
    &$86.03{\scriptstyle \pm 0.58}$
    \\
    IEPT~\cite{zhang2020iept}
    &ResNet-12
    &12.4 M
    &$67.05{\scriptstyle \pm 0.44}$
    &$82.90{\scriptstyle \pm 0.30}$
    &$72.24{\scriptstyle \pm 0.50}$
    &$86.73{\scriptstyle \pm 0.34}$
    \\
    MELR~\cite{fei2020melr}
    &ResNet-12
    &12.4 M
    &$67.40{\scriptstyle \pm 0.43}$
    &$83.40{\scriptstyle \pm 0.28}$
    &$72.14{\scriptstyle \pm 0.51}$
    &$87.01{\scriptstyle \pm 0.35}$
    \\
    FRN~\cite{Wertheimer_2021_CVPR}
    &ResNet-12
    &12.4 M
    &$66.45{\scriptstyle \pm 0.19}$
    &$82.83{\scriptstyle \pm 0.13}$
    &$72.06{\scriptstyle \pm 0.22}$
    &$86.89{\scriptstyle \pm 0.14}$
    \\
    CG~\cite{zhao2021looking}
    &ResNet-12
    &12.4 M
    &$67.02{\scriptstyle \pm 0.20}$
    &$82.32{\scriptstyle \pm 0.14}$
    &$71.66{\scriptstyle \pm 0.23}$
    &$85.50{\scriptstyle \pm 0.15}$
    \\
    DMF~\cite{xu2021learning}
    &ResNet-12
    &12.4 M
    &$67.76{\scriptstyle \pm 0.46}$
    &$82.71{\scriptstyle \pm 0.31}$
    &$71.89{\scriptstyle \pm 0.52}$
    &$85.96{\scriptstyle \pm 0.35}$
    \\
    InfoPatch~\cite{liu2021learning}
    &ResNet-12
    &12.4 M
    &$67.67{\scriptstyle \pm 0.45}$
    &$82.44{\scriptstyle \pm 0.31}$
    &-
    &-
    \\
    BML~\cite{zhou2021binocular}
    &ResNet-12
    &12.4 M
    &$67.04{\scriptstyle \pm 0.63}$
    &$83.63{\scriptstyle \pm 0.29}$
    &$68.99{\scriptstyle \pm 0.50}$
    &$85.49{\scriptstyle \pm 0.34}$
    \\
    CNL~\cite{zhao2021looking}
    &ResNet-12
    &12.4 M
    &$67.96{\scriptstyle \pm 0.98}$
    &$83.36{\scriptstyle \pm 0.51}$
    &$73.42{\scriptstyle \pm 0.95}$
    &$87.72{\scriptstyle \pm 0.75}$
    \\
    Meta-NVG~\cite{zhang2021meta}
    &ResNet-12
    &12.4 M
    &$67.14{\scriptstyle \pm 0.80}$
    &$83.82{\scriptstyle \pm 0.51}$
    &$74.58{\scriptstyle \pm 0.88}$
    &$86.73{\scriptstyle \pm 0.61}$
    \\
    PAL~\cite{ma2021partner}
    &ResNet-12
    &12.4 M
    &$69.37{\scriptstyle \pm 0.64}$
    &$84.40{\scriptstyle \pm 0.44}$
    &$72.25{\scriptstyle \pm 0.72}$
    &$86.95{\scriptstyle \pm 0.47}$
    \\
    COSOC~\cite{luo2021rectifying}
    &ResNet-12
    &12.4 M
    &$69.28{\scriptstyle \pm 0.49}$
    &$85.16{\scriptstyle \pm 0.42}$
    &$73.57{\scriptstyle \pm 0.43}$
    &$87.57{\scriptstyle \pm 0.10}$
    \\
    Meta DeepBDC~\cite{DeepBDC-CVPR2022}
    &ResNet-12
    &12.4 M
    &$67.34{\scriptstyle \pm 0.43}$
    &$84.46{\scriptstyle \pm 0.28}$
    &$72.34{\scriptstyle \pm 0.49}$
    &$87.31{\scriptstyle \pm 0.32}$
    \\
    \hline 
    LEO~\cite{rusu2018meta}
    &WRN-28-10
    &36.5 M
    &$61.76{\scriptstyle \pm 0.08}$
    &$77.59{\scriptstyle \pm 0.12}$
    &$66.33{\scriptstyle \pm 0.05}$
    &$81.44{\scriptstyle \pm 0.09}$
    \\
    CC+rot~\cite{gidaris2019boosting}
    &WRN-28-10
    &36.5 M
    &$62.93{\scriptstyle \pm 0.45}$
    &$79.87{\scriptstyle \pm 0.33}$
    &$70.53{\scriptstyle \pm 0.51}$
    &$84.98{\scriptstyle \pm 0.36}$
    \\
    FEAT~\cite{ye2020fewshot}
    &WRN-28-10
    &36.5 M
    &$65.10{\scriptstyle \pm 0.20}$
    &$81.11{\scriptstyle \pm 0.14}$
    &$70.41{\scriptstyle \pm 0.23}$
    &$84.38{\scriptstyle \pm 0.16}$
    \\
    PSST~\cite{chen2021pareto}
    &WRN-28-10
    &36.5 M
    &$64.16{\scriptstyle \pm 0.44}$
    &$80.64{\scriptstyle \pm 0.32}$
    &-
    &-
    \\
    MetaQDA~\cite{zhang2021shallow}
    &WRN-28-10
    &36.5 M
    &$67.83{\scriptstyle \pm 0.64}$
    &$84.28{\scriptstyle \pm 0.69}$
    &$74.33{\scriptstyle \pm 0.65}$
    &$89.56{\scriptstyle \pm 0.79}$
    \\
    OM~\cite{qi2021transductive}
    &WRN-28-10
    &36.5 M
    &$66.78{\scriptstyle \pm 0.30}$
    &$85.29{\scriptstyle \pm 0.41}$
    &$71.54{\scriptstyle \pm 0.29}$
    &$87.79{\scriptstyle \pm 0.46}$
    
    \\
    \hline
    
    \rowcolor{Apricot!20!White} FewTURE (ours)
    % \rowcolor{Gray!70!White} FewTURE (ours)
    &ViT-Small
    &22 M
    &$68.02{\scriptstyle \pm 0.88}$
    &$84.51{\scriptstyle \pm 0.53}$
    &$72.96{\scriptstyle \pm 0.92}$
    &$86.43{\scriptstyle \pm 0.67}$
    \\
    % \rowcolor{Apricot!30!White} FewTURE (ours)
    % &Swin-Tiny
    \rowcolor{Apricot!20!White} FewTURE (ours)
    &Swin-Tiny
    &29 M
    &$\bf{72.40{\scriptstyle \pm 0.78}}$
    &$\bf{86.38{\scriptstyle \pm 0.49}}$
    &$\bf{76.32{\scriptstyle \pm 0.87}}$
    &$\bf{89.96{\scriptstyle \pm 0.55}}$
    \\
    \hline
    \end{tabular}
    }
    \label{tab:benchmark_mini_tiered}
\end{table}

\begin{table}[tb]
    % \vspace{-1.5em}
    \centering
    \caption{Average classification accuracy for $5$-way $1$-shot and $5$-way $5$-shot scenarios. Reported are the mean and 95\% confidence interval on the unseen test sets of CIFAR-FS~\cite{bertinetto2019_cifarfsl} and FC-100~\cite{oreshkin2018_tadam_fc100}, using the established evaluation protocols.}
    \scalebox{0.85}
    {
    \begin{tabular}{l|c c|c c|c c}
    \specialrule{.2em}{.1em}{.1em}
    \multirow{2}{*}{\textbf{Model}}
    &\multirow{2}{*}{\textbf{Backbone}}
    &\multirow{2}{*}{\textbf{$\approx$~\# Params}}
    &\multicolumn{2}{c|}{\textbf{CIFAR-FS}}
    &\multicolumn{2}{c}{\textbf{FC100}}
    \\
    &
    &
    &\textbf{1-shot}
    &\textbf{5-shot}
    &\textbf{1-shot}
    &\textbf{5-shot}
    \\
    \toprule\toprule
    ProtoNet~\cite{snell2017prototypical}
    &ResNet-12
    &12.4 M
    &-
    &-
    &$41.54{\scriptstyle \pm 0.76}$
    &$57.08{\scriptstyle \pm 0.76}$
    \\
    MetaOpt~\cite{lee2019meta}
    &ResNet-12
    &12.4 M
    &$72.00{\scriptstyle \pm 0.70}$
    &$84.20{\scriptstyle \pm 0.50}$
    &$41.10{\scriptstyle \pm 0.60}$
    &$55.50{\scriptstyle \pm 0.60}$
    \\
    MABAS~\cite{kim2020_mabas}
    &ResNet-12
    &12.4 M
    &$73.51{\scriptstyle \pm 0.92}$
    &$85.65{\scriptstyle \pm 0.65}$
    &$42.31{\scriptstyle \pm 0.75}$
    &$58.16{\scriptstyle \pm 0.78}$
    \\
    RFS~\cite{tian2020rethinking}
    &ResNet-12
    &12.4 M
    &$73.90{\scriptstyle \pm 0.80}$
    &$86.90{\scriptstyle \pm 0.50}$
    &$44.60{\scriptstyle \pm 0.70}$
    &$60.90{\scriptstyle \pm 0.60}$
    \\
    BML~\cite{zhou2021binocular}
    &ResNet-12
    &12.4 M
    &$73.45{\scriptstyle \pm 0.47}$
    &$88.04{\scriptstyle \pm 0.33}$
    &-
    &-
    \\
    CG~\cite{gao2021curvature}
    &ResNet-12
    &12.4 M
    &$73.00{\scriptstyle \pm 0.70}$
    &$85.80{\scriptstyle \pm 0.50}$
    &-
    &-
    \\
    Meta-NVG~\cite{zhang2021meta}
    &ResNet-12
    &12.4 M
    &$74.63{\scriptstyle \pm 0.91}$
    &$86.45{\scriptstyle \pm 0.59}$
    &$46.40{\scriptstyle \pm 0.81}$
    &$61.33{\scriptstyle \pm 0.71}$
    \\
    RENet~\cite{kang2021renet}
    &ResNet-12
    &12.4 M
    &$74.51{\scriptstyle \pm 0.46}$
    &$86.60{\scriptstyle \pm 0.32}$
    &-
    &-
    \\
    TPMN~\cite{wu2021task}
    &ResNet-12
    &12.4 M
    &$75.50{\scriptstyle \pm 0.90}$
    &$87.20{\scriptstyle \pm 0.60}$
    &$46.93{\scriptstyle \pm 0.71}$
    &$63.26{\scriptstyle \pm 0.74}$
    \\
    MixFSL~\cite{Afrasiyabi_2021_ICCV}
    &ResNet-12
    &12.4 M
    &-
    &-
    &$44.89{\scriptstyle \pm 0.63}$
    &$60.70{\scriptstyle \pm 0.60}$
    \\
    \hline
    CC+rot~\cite{gidaris2019boosting}
    &WRN-28-10
    &36.5 M
    &$73.62{\scriptstyle \pm 0.31}$
    &$86.05{\scriptstyle \pm 0.22}$
    &-
    &-
    \\
    PSST~\cite{chen2021pareto}
    &WRN-28-10
    &36.5 M
    &$77.02{\scriptstyle \pm 0.38}$
    &$88.45{\scriptstyle \pm 0.35}$
    &-
    &-
    \\
    Meta-QDA~\cite{zhang2021shallow}
    &WRN-28-10
    &36.5 M
    &$75.83{\scriptstyle \pm 0.88}$
    &$88.79{\scriptstyle \pm 0.75}$
    &-
    &-
    
    \\
    \hline
    \rowcolor{Apricot!20!White} FewTURE (ours)
    &ViT-Small
    &22 M
    &$76.10{\scriptstyle \pm 0.88}$
    &$86.14{\scriptstyle \pm 0.64}$
    &$46.20{\scriptstyle \pm 0.79}$
    &$63.14{\scriptstyle \pm 0.73}$
    \\
    \rowcolor{Apricot!20!White} FewTURE (ours)
    &Swin-Tiny
    &29 M
    &$\bf{77.76{\scriptstyle \pm 0.81}}$
    &$\bf{88.90{\scriptstyle \pm 0.59}}$
    &$\bf{47.68{\scriptstyle \pm 0.78}}$
    &$\bf{63.81{\scriptstyle \pm 0.75}}$
    \\
    \hline
    \end{tabular}
    }
    \label{tab:benchmark_cifarfs_fc100}
\end{table}
\subsection{Evaluation on few-shot classification benchmarks}
We conduct experiments using the few-shot settings established in the community, namely $5$-way $1$-shot and $5$-way $5$-shot -- meaning the network has to distinguish samples from 5 novel classes based on a provided number of 1 or 5 images per class. We evaluate our method \textit{FewTURE} using two different Transformer backbones and compare our results against the current state of the art in~\cref{tab:benchmark_mini_tiered} for the \textit{mini}ImageNet and \textit{tiered}ImageNet, and in~\cref{tab:benchmark_cifarfs_fc100} for the CIFAR-FS and FC100 datasets. It is to be noted that in contrast to previous works, we do not employ the help of any convolutional backbone but instead (and as far as we are aware for the first time) use a Transformer backbone together with our previously introduced token importance reweighting method to achieve these results. We are able to set new state of the art results across all four datasets in both $5$-shot and $1$-shot settings, improving particularly the $1$-shot results on \textit{mini}ImageNet and \textit{tiered}ImageNet by significant margins of $3.03\%$ and $1.74\%$, respectively.

\begin{wraptable}{R}{4.5cm}
\vspace{-1.3em}
\caption{Pruning the number of tokens. Test accuracy for $5$-way $5$-shot on \textit{mini}ImageNet~\cite{vinyals2016_miniimagenet}.}\label{tab:tokenpruning}
\centering
\begin{tabular}{rc}
\specialrule{.2em}{.1em}{.1em}
\textbf{\# tokens} & \textbf{Test Acc.}\\
\toprule\toprule
$100\%$ & $84.05\pm0.53$ \\  
$75\%$ & $83.15\pm0.57$ \\  
$50\%$ & $83.81\pm0.59$ \\ 
$25\%$ & $81.79\pm0.57$ \\
$10\%$ & $81.05\pm0.62$ \\
\bottomrule
\end{tabular}
\end{wraptable} 
\subsection{Increasing efficiency by pruning the token sequence}
To further improve the computational efficiency of our method, we investigate its behaviour when limiting the number of tokens that are considered to establish patch-wise correspondences to only a subset -- allowing \textit{FewTURE} to scale to potentially large \textit{many-way many-shot} settings. We use the attention maps inherent in our approach (averaged over all heads) to prune the number of patch tokens by only using the ones within the top-k attention values to compute the similarity matrix~$\boldsymbol{S}$. The results obtained with our ViT-small backbone for pruning the number of tokens to $75\%$, $50\%$, $25\%$ and $10\%$ of the original token number indicate that such pruning might be an interesting avenue for future work, 
with our method still achieving $96.4\%$ of its original performance when only retaining $10\%$ of the number of tokens.

\begin{wraptable}{L}{6.5cm}
\vspace{-1.3em}
\caption{Changing the classifier. Test accuracy for $5$-way $5$-shot on \textit{mini}ImageNet~\cite{vinyals2016_miniimagenet}.}\label{tab:abl_classifier}
\centering
\begin{tabular}{lc}
\specialrule{.2em}{.1em}{.1em}
\textbf{Classifier} & \textbf{Test Acc.}\\
\toprule\toprule
Prototyp. w/ Euclid. Dist. & $82.80\pm0.59$ \\  
Prototyp. w/ Cosine. Dist. & $79.90\pm0.65$ \\  
Linear (optimized online) & $82.37\pm0.57$ \\ 
FewTURE (~0 rew. steps)  & $82.68\pm0.55$ \\
FewTURE (15 rew. steps) & $\mathbf{84.05}\pm\mathbf{0.53}$ \\
\bottomrule
\end{tabular}
\end{wraptable} 
\subsection{Ablation on the type of classifier}
We train a linear classifier as well as prototypical approach using our pre-trained ViT-small backbone for a 5-way 5-shot setting on \textit{mini}Imagenet~\cite{vinyals2016_miniimagenet} to investigate the influence of the choice of classifier (Table~\ref{tab:abl_classifier}). 
We were able to obtain a test accuracy of $82.80\%$ for the prototypical network after optimising the pre-trained backbone with meta-finetuning, which is competitive to the results we obtain for our method without reweighting (’0 step’ in Figure~\ref{fig:ablation_upt_steps_both}(\subref{subfig:ablation_upt_steps_testacc})), but is clearly outperformed by our reweighting-based approach. To provide a fair comparison, we optimize the linear classifier at inference time to adapt to the support set and obtained a maximum test accuracy of $82.37\%$.  Both results indicate the quality of embedding our backbone is able to produce but also demonstrate the importance of our task-specific reweighting-based approach. For further ablation studies, please refer to the supplementary material.

\subsection{Limitations} \label{sec:limits}
Our introduced method is arguably more powerful for cases where multiple examples of the same class are provided, \ie in $N$-way $K$-shot settings with $K>1$. While FewTURE works well in $1$-shot settings (\cref{tab:benchmark_mini_tiered} and \ref{tab:benchmark_cifarfs_fc100}), \ignore{it is to be noted that }the inner loop adaptation procedure still aims to exclude cross-class similarities that hurt classification performance, but has less diverse information for selecting the most helpful regions due to the lack of other in-class comparison samples -- which might thus yield slightly less-refined token selections compared to multi-shot scenarios (\textit{supplementary material}). 

While we did not face significant problems regarding the comparably very small size of our datasets (\eg \textit{mini}ImageNet with 38K compared to the usually used ImageNet with 1.28M training images), more specialized applications with highly limited training data might be negatively impacted and successfully training our method due to the reduced inductive bias present in the Transformer architecture could prove challenging.

\section{Related work}
Over the past few years, the family of few-shot learning (FSL) has grown diverse and broad. Those closely related to this work can be categorized into two groups: metric-based methods~\cite{ma2021learning,ma2021adaptive,simon2020adaptive,  snell2017prototypical,vinyals2016_miniimagenet,ye2020fewshot,Zhang_2020_CVPR} and optimisation-based methods~\cite{finn2017model,lee2019meta,nichol2018first,rusu2018meta, zintgraf2019fast}. 
Metric-based methods, such as ProtoNet~\cite{snell2017prototypical}, DeepEMD~\cite{Zhang_2020_CVPR}, and RelationNet~\cite{sung2018learning} aim to learn a class representation (prototype) by averaging the embeddings belonging to the same class and employ a predefined (ProtoNet and DeepEMD) or learned (RelationNet) metric to perform prototype-query matching. FEAT~\cite{ye2020fewshot} and TDM~\cite{lee2022task} take this a step further and use attention mechanisms to adapt the extracted features to the novel tasks\ignore{, resulting in a more adaptive method}. Our method instead fully utilizes the embeddings of local image regions (patch tokens)\ignore{encoded by the feature extractor}, preventing loss of information and supervision collapse occurring in the aforementioned prototype-based approaches (see supplementary material). 

Optimisation-based methods such as MAML~\cite{finn2017model} and Reptile~\cite{nichol2018first} propose to learn a set of initial model parameters that can quickly adapt to a novel task. However, updating all model parameters is often not feasible given large backbones and only few labeled samples during inference. To alleviate this so-called meta-overfitting problem, CAVIA~\cite{zintgraf2019fast} and LEO~\cite{rusu2018meta} propose to learn and adapt a lower dimensional representation that is mapped onto the network. Our method is inspired by such lower-dimensional adaptation strategies and learns a tiny set of context-aware re-weighting factors online for each novel task without requiring higher-order gradients, resulting in a flexible and efficient framework.  

\textbf{Self-supervised learning for FSL}. Although self-supervised learning is underrepresented in the context of few-shot learning, some recent works~\cite{gidaris2019boosting, mangla2020charting, su2020does} have shown that self-supervision via pretext tasks can be beneficial when integrated as auxiliary loss. S2M2~\cite{mangla2020charting} employs rotation~\cite{gidaris2018unsupervised} and exemplars~\cite{dosovitskiy2014discriminative} along with common supervised learning during the pre-training stage. CTX~\cite{doersch2020_crosstransformers} demonstrates that SimCLR~\cite{chen2020simple} can be combined with supervised-learning tasks in an episodic-training manner to learn a more generalized model. In contrast, FewTURE demonstrates that the self-supervised pretext task (\ie, Masked Image Modelling ~\cite{atito2021sit,bao2021_beit,li2021mst,he2021masked,pathak2016context,tan2021vimpac,zhou2021_ibot}), can be used in a standalone manner to learn more generalized features for few-shot learning on small-scale datasets.

\par
\textbf{Vision Transformers in FSL}.
Transformers have been immensely successful in the field of Natural Language Processing. Recent studies~\cite{dosovitskiy2020_vit,li2021efficient,liu2021_swin,tu2022maxvit} suggest that encoding the long-range dependency of data via self-attention also yields promising results for vision tasks (\eg, image classification, joint vision-language modeling, \etc). However, there is an important trade-off between leveraging the rich representation capacity and the lack of inductive bias. Transformers have gained a reputation to generally require significantly more training data compared to convolutional neural networks (CNNs), since properties like translation invariance, locality and hierarchical structure of visual data have to be inferred from the data~\cite{liu2021efficient}. While this data-hungry nature largely prevented the use of Transformers in problems with scarce data like few-shot learning, some recent works demonstrate that a single Transformer head can be successfully adopted to perform feature adaptation~\cite{doersch2020_crosstransformers,ye2020fewshot}. To the best of our knowledge, \textit{FewTURE} is the first approach that uses a fully Transformer-based architecture to obtain representative embeddings while being trained exclusively on training data of the respective few-shot dataset.

\section{Conclusion}
%One-hot image-level annotations sometimes only describe a subset of the image content. This might be acceptable for normal image classification whereas harmful for few-shot scenario, where the supervision collapse could potentially lead to classification failure given the categorical discrepancy between training and testing. 
In this paper, we presented a novel approach to tackle the challenge of supervision collapse introduced by one-hot image-level labels in few-shot learning. We split the input images into smaller parts with higher probability of being dominated by only one entity and encode these local regions using a Vision Transformer architecture pretrained via Masked Image Modeling (MIM) in a self-supervised way to learn a representative embedding space beyond the pure label information. We devise a classifier based on patch embedding similarities and propose a token importance reweighting mechanism to refine the contribution of each local patch towards the overall classification result based around intra-class similarities and inter-class differences as a function of the support set information. Our obtained results demonstrate that our proposed method alleviates the problem of supervision collapse by learning more generalized features, achieving new state-of-the-art results on four popular few-shot classification datasets.\par
\noindent \textbf{Acknowledgements.} The authors would like to thank Zhou~\textit{et al.}~\cite{zhou2021_ibot} for sharing their insights and code regarding self-supervised pretraining, as well as Dosovitskiy~\textit{et al.}~\cite{dosovitskiy2020_vit}, Touvron~\textit{et al.}~\cite{touvron2021_deit} and Liu~\textit{et al.}~\cite{liu2021_swin} for sharing details of the ViT and Swin architectures. \par
\noindent Parts of this research were undertaken using the LIEF HPC-GPGPU Facility hosted at the University of Melbourne. This Facility was established with the assistance of LIEF Grant LE170100200. 

{
\small
\bibliographystyle{ieee_fullname}
\bibliography{patchfs}

\begin{thebibliography}{10}\itemsep=-1pt

\bibitem{Afrasiyabi_2021_ICCV}
Arman Afrasiyabi, Jean-Fran{\c{c}}ois Lalonde, and Christian Gagn{\'e}.
\newblock Mixture-based feature space learning for few-shot image
  classification.
\newblock In {\em Proceedings of the IEEE/CVF International Conference on
  Computer Vision (ICCV)}, pages 9041--9051, 2021.

\bibitem{atito2021sit}
Sara Atito, Muhammad Awais, and Josef Kittler.
\newblock Sit: Self-supervised vision transformer.
\newblock {\em arXiv preprint arXiv:2104.03602}, 2021.

\bibitem{bao2021_beit}
Hangbo Bao, Li Dong, Songhao Piao, and Furu Wei.
\newblock {BE}it: {BERT} pre-training of image transformers.
\newblock In {\em International Conference on Learning Representations}, 2022.

\bibitem{bertinetto2019_cifarfsl}
Luca Bertinetto, Joao~F Henriques, Philip Torr, and Andrea Vedaldi.
\newblock Meta-learning with differentiable closed-form solvers.
\newblock In {\em International Conference on Learning Representations}, 2019.

\bibitem{caron2021_dino}
Mathilde Caron, Hugo Touvron, Ishan Misra, Herv{\'e} J{\'e}gou, Julien Mairal,
  Piotr Bojanowski, and Armand Joulin.
\newblock Emerging properties in self-supervised vision transformers.
\newblock In {\em Proceedings of the IEEE/CVF International Conference on
  Computer Vision}, pages 9650--9660, 2021.

\bibitem{chen2020simple}
Ting Chen, Simon Kornblith, Mohammad Norouzi, and Geoffrey Hinton.
\newblock A simple framework for contrastive learning of visual
  representations.
\newblock In {\em International Conference on Machine Learning}, pages
  1597--1607. PMLR, 2020.

\bibitem{chen2021_mocov3}
Xinlei Chen, Saining Xie, and Kaiming He.
\newblock An empirical study of training self-supervised vision transformers.
\newblock In {\em Proceedings of the IEEE/CVF International Conference on
  Computer Vision}, pages 9640--9649, 2021.

\bibitem{chen2021pareto}
Zhengyu Chen, Jixie Ge, Heshen Zhan, Siteng Huang, and Donglin Wang.
\newblock Pareto self-supervised training for few-shot learning.
\newblock In {\em Proceedings of the IEEE/CVF Conference on Computer Vision and
  Pattern Recognition}, pages 13663--13672, 2021.

\bibitem{doersch2020_crosstransformers}
Carl Doersch, Ankush Gupta, and Andrew Zisserman.
\newblock Crosstransformers: spatially-aware few-shot transfer.
\newblock {\em Advances in Neural Information Processing Systems},
  33:21981--21993, 2020.

\bibitem{dosovitskiy2020_vit}
Alexey Dosovitskiy, Lucas Beyer, Alexander Kolesnikov, Dirk Weissenborn,
  Xiaohua Zhai, Thomas Unterthiner, Mostafa Dehghani, Matthias Minderer, Georg
  Heigold, Sylvain Gelly, et~al.
\newblock An image is worth 16x16 words: Transformers for image recognition at
  scale.
\newblock {\em arXiv preprint arXiv:2010.11929}, 2020.

\bibitem{dosovitskiy2014discriminative}
Alexey Dosovitskiy, Jost~Tobias Springenberg, Martin Riedmiller, and Thomas
  Brox.
\newblock Discriminative unsupervised feature learning with convolutional
  neural networks.
\newblock {\em Advances in Neural Information Processing Systems}, 27, 2014.

\bibitem{fei2020melr}
Nanyi Fei, Zhiwu Lu, Tao Xiang, and Songfang Huang.
\newblock Melr: Meta-learning via modeling episode-level relationships for
  few-shot learning.
\newblock In {\em International Conference on Learning Representations}, 2020.

\bibitem{finn2017model}
Chelsea Finn, Pieter Abbeel, and Sergey Levine.
\newblock Model-agnostic meta-learning for fast adaptation of deep networks.
\newblock In {\em International Conference on Machine Learning}, pages
  1126--1135. PMLR, 2017.

\bibitem{gao2021curvature}
Zhi Gao, Yuwei Wu, Yunde Jia, and Mehrtash Harandi.
\newblock Curvature generation in curved spaces for few-shot learning.
\newblock In {\em Proceedings of the IEEE/CVF International Conference on
  Computer Vision}, pages 8691--8700, 2021.

\bibitem{gidaris2019boosting}
Spyros Gidaris, Andrei Bursuc, Nikos Komodakis, Patrick P{\'e}rez, and Matthieu
  Cord.
\newblock Boosting few-shot visual learning with self-supervision.
\newblock In {\em Proceedings of the IEEE/CVF International Conference on
  Computer Vision}, pages 8059--8068, 2019.

\bibitem{gidaris2018unsupervised}
Spyros Gidaris, Praveer Singh, and Nikos Komodakis.
\newblock Unsupervised representation learning by predicting image rotations.
\newblock In {\em International Conference on Learning Representations}, 2018.

\bibitem{he2021masked}
Kaiming He, Xinlei Chen, Saining Xie, Yanghao Li, Piotr Doll{\'a}r, and Ross
  Girshick.
\newblock Masked autoencoders are scalable vision learners.
\newblock {\em arXiv preprint arXiv:2111.06377}, 2021.

\bibitem{hou2019_crossattention}
Ruibing Hou, Hong Chang, Bingpeng Ma, Shiguang Shan, and Xilin Chen.
\newblock Cross attention network for few-shot classification.
\newblock {\em Advances in Neural Information Processing Systems}, 32, 2019.

\bibitem{kang2021renet}
Dahyun Kang, Heeseung Kwon, Juhong Min, and Minsu Cho.
\newblock Relational embedding for few-shot classification.
\newblock In {\em Proceedings of the IEEE/CVF International Conference on
  Computer Vision (ICCV)}, 2021.

\bibitem{kim2020_mabas}
Jaekyeom Kim, Hyoungseok Kim, and Gunhee Kim.
\newblock Model-agnostic boundary-adversarial sampling for test-time
  generalization in few-shot learning.
\newblock In {\em Computer Vision--ECCV 2020: 16th European Conference,
  Glasgow, UK, August 23--28, 2020, Proceedings, Part I 16}, pages 599--617.
  Springer, 2020.

\bibitem{lee2019meta}
Kwonjoon Lee, Subhransu Maji, Avinash Ravichandran, and Stefano Soatto.
\newblock Meta-learning with differentiable convex optimization.
\newblock In {\em Proceedings of the IEEE/CVF Conference on Computer Vision and
  Pattern Recognition}, pages 10657--10665, 2019.

\bibitem{lee2022task}
SuBeen Lee, WonJun Moon, and Jae-Pil Heo.
\newblock Task discrepancy maximization for fine-grained few-shot
  classification.
\newblock In {\em Proceedings of the IEEE/CVF Conference on Computer Vision and
  Pattern Recognition}, pages 5331--5340, 2022.

\bibitem{li2021efficient}
Chunyuan Li, Jianwei Yang, Pengchuan Zhang, Mei Gao, Bin Xiao, Xiyang Dai, Lu
  Yuan, and Jianfeng Gao.
\newblock Efficient self-supervised vision transformers for representation
  learning.
\newblock {\em International Conference on Learning Representations (ICLR)},
  2022.

\bibitem{li2021mst}
Zhaowen Li, Zhiyang Chen, Fan Yang, Wei Li, Yousong Zhu, Chaoyang Zhao, Rui
  Deng, Liwei Wu, Rui Zhao, Ming Tang, et~al.
\newblock Mst: Masked self-supervised transformer for visual representation.
\newblock {\em Advances in Neural Information Processing Systems}, 34, 2021.

\bibitem{liu2021learning}
Chen Liu, Yanwei Fu, Chengming Xu, Siqian Yang, Jilin Li, Chengjie Wang, and Li
  Zhang.
\newblock Learning a few-shot embedding model with contrastive learning.
\newblock In {\em Proceedings of the AAAI Conference on Artificial
  Intelligence}, volume~35, pages 8635--8643, 2021.

\bibitem{liu2021efficient}
Yahui Liu, Enver Sangineto, Wei Bi, Nicu Sebe, Bruno Lepri, and Marco Nadai.
\newblock Efficient training of visual transformers with small datasets.
\newblock {\em Advances in Neural Information Processing Systems}, 34, 2021.

\bibitem{liu2021_swin}
Ze Liu, Yutong Lin, Yue Cao, Han Hu, Yixuan Wei, Zheng Zhang, Stephen Lin, and
  Baining Guo.
\newblock Swin transformer: Hierarchical vision transformer using shifted
  windows.
\newblock In {\em Proceedings of the IEEE/CVF International Conference on
  Computer Vision}, pages 10012--10022, 2021.

\bibitem{luo2021rectifying}
Xu Luo, Longhui Wei, Liangjian Wen, Jinrong Yang, Lingxi Xie, Zenglin Xu, and
  Qi Tian.
\newblock Rectifying the shortcut learning of background for few-shot learning.
\newblock {\em Advances in Neural Information Processing Systems}, 34, 2021.

\bibitem{ma2021partner}
Jiawei Ma, Hanchen Xie, Guangxing Han, Shih-Fu Chang, Aram Galstyan, and Wael
  Abd-Almageed.
\newblock Partner-assisted learning for few-shot image classification.
\newblock In {\em Proceedings of the IEEE/CVF International Conference on
  Computer Vision}, pages 10573--10582, 2021.

\bibitem{ma2021learning}
Rongkai Ma, Pengfei Fang, Gil Avraham, Yan Zuo, Tom Drummond, and Mehrtash
  Harandi.
\newblock Learning instance and task-aware dynamic kernels for few shot
  learning.
\newblock {\em arXiv preprint arXiv:2112.03494}, 2021.

\bibitem{ma2021adaptive}
Rongkai Ma, Pengfei Fang, Tom Drummond, and Mehrtash Harandi.
\newblock Adaptive poincar\'e point to set distance for few-shot
  classification.
\newblock {\em arXiv preprint arXiv:2112.01719}, 2021.

\bibitem{mangla2020charting}
Puneet Mangla, Nupur Kumari, Abhishek Sinha, Mayank Singh, Balaji
  Krishnamurthy, and Vineeth~N Balasubramanian.
\newblock Charting the right manifold: Manifold mixup for few-shot learning.
\newblock In {\em Proceedings of the IEEE/CVF Winter Conference on Applications
  of Computer Vision}, pages 2218--2227, 2020.

\bibitem{nichol2018first}
Alex Nichol, Joshua Achiam, and John Schulman.
\newblock On first-order meta-learning algorithms.
\newblock {\em arXiv preprint arXiv:1803.02999}, 2018.

\bibitem{oreshkin2018_tadam_fc100}
Boris Oreshkin, Pau Rodr{\'\i}guez~L{\'o}pez, and Alexandre Lacoste.
\newblock Tadam: Task dependent adaptive metric for improved few-shot learning.
\newblock {\em Advances in Neural Information Processing Systems}, 31, 2018.

\bibitem{pathak2016context}
Deepak Pathak, Philipp Krahenbuhl, Jeff Donahue, Trevor Darrell, and Alexei~A
  Efros.
\newblock Context encoders: Feature learning by inpainting.
\newblock In {\em Proceedings of the IEEE Conference on Computer Vision and
  Pattern Recognition}, pages 2536--2544, 2016.

\bibitem{qi2021transductive}
Guodong Qi, Huimin Yu, Zhaohui Lu, and Shuzhao Li.
\newblock Transductive few-shot classification on the oblique manifold.
\newblock In {\em Proceedings of the IEEE/CVF International Conference on
  Computer Vision}, pages 8412--8422, 2021.

\bibitem{ren2018_tieredimagenet}
Mengye Ren, Eleni Triantafillou, Sachin Ravi, Jake Snell, Kevin Swersky,
  Joshua~B Tenenbaum, Hugo Larochelle, and Richard~S Zemel.
\newblock Meta-learning for semi-supervised few-shot classification.
\newblock {\em arXiv preprint arXiv:1803.00676}, 2018.

\bibitem{russakovsky_2015imagenet}
Olga Russakovsky, Jia Deng, Hao Su, Jonathan Krause, Sanjeev Satheesh, Sean Ma,
  Zhiheng Huang, Andrej Karpathy, Aditya Khosla, Michael Bernstein, et~al.
\newblock Imagenet large scale visual recognition challenge.
\newblock {\em International Journal of Computer Vision}, 115(3):211--252,
  2015.

\bibitem{rusu2018meta}
Andrei~A Rusu, Dushyant Rao, Jakub Sygnowski, Oriol Vinyals, Razvan Pascanu,
  Simon Osindero, and Raia Hadsell.
\newblock Meta-learning with latent embedding optimization.
\newblock In {\em International Conference on Learning Representations}, 2018.

\bibitem{simon2020adaptive}
Christian Simon, Piotr Koniusz, Richard Nock, and Mehrtash Harandi.
\newblock Adaptive subspaces for few-shot learning.
\newblock In {\em Proceedings of the IEEE/CVF Conference on Computer Vision and
  Pattern Recognition}, pages 4136--4145, 2020.

\bibitem{snell2017prototypical}
Jake Snell, Kevin Swersky, and Richard Zemel.
\newblock Prototypical networks for few-shot learning.
\newblock {\em Advances in Neural Information Processing Systems}, 30, 2017.

\bibitem{su2020does}
Jong-Chyi Su, Subhransu Maji, and Bharath Hariharan.
\newblock When does self-supervision improve few-shot learning?
\newblock In {\em European Conference on Computer Vision}, pages 645--666.
  Springer, 2020.

\bibitem{sung2018learning}
Flood Sung, Yongxin Yang, Li Zhang, Tao Xiang, Philip~HS Torr, and Timothy~M
  Hospedales.
\newblock Learning to compare: Relation network for few-shot learning.
\newblock In {\em Proceedings of the IEEE Conference on Computer Vision and
  Pattern Recognition}, pages 1199--1208, 2018.

\bibitem{tan2021vimpac}
Hao Tan, Jie Lei, Thomas Wolf, and Mohit Bansal.
\newblock Vimpac: Video pre-training via masked token prediction and
  contrastive learning.
\newblock {\em arXiv preprint arXiv:2106.11250}, 2021.

\bibitem{tian2020rethinking}
Yonglong Tian, Yue Wang, Dilip Krishnan, Joshua~B Tenenbaum, and Phillip Isola.
\newblock Rethinking few-shot image classification: a good embedding is all you
  need?
\newblock In {\em European Conference on Computer Vision}, pages 266--282.
  Springer, 2020.

\bibitem{touvron2021_deit}
Hugo Touvron, Matthieu Cord, Matthijs Douze, Francisco Massa, Alexandre
  Sablayrolles, and Herv{\'e} J{\'e}gou.
\newblock Training data-efficient image transformers \& distillation through
  attention.
\newblock In {\em International Conference on Machine Learning}, pages
  10347--10357. PMLR, 2021.

\bibitem{tu2022maxvit}
Zhengzhong Tu, Hossein Talebi, Han Zhang, Feng Yang, Peyman Milanfar, Alan
  Bovik, and Yinxiao Li.
\newblock Maxvit: Multi-axis vision transformer.
\newblock {\em arXiv preprint arXiv:2204.01697}, 2022.

\bibitem{vinyals2016_miniimagenet}
Oriol Vinyals, Charles Blundell, Timothy Lillicrap, Daan Wierstra, et~al.
\newblock Matching networks for one shot learning.
\newblock {\em Advances in Neural Information Processing Systems},
  29:3630--3638, 2016.

\bibitem{Wertheimer_2021_CVPR}
Davis Wertheimer, Luming Tang, and Bharath Hariharan.
\newblock Few-shot classification with feature map reconstruction networks.
\newblock In {\em Proceedings of the IEEE/CVF Conference on Computer Vision and
  Pattern Recognition (CVPR)}, pages 8012--8021, 2021.

\bibitem{wu2021task}
Jiamin Wu, Tianzhu Zhang, Yongdong Zhang, and Feng Wu.
\newblock Task-aware part mining network for few-shot learning.
\newblock In {\em Proceedings of the IEEE/CVF International Conference on
  Computer Vision}, pages 8433--8442, 2021.

\bibitem{DeepBDC-CVPR2022}
Jiangtao Xie, Fei Long, Jiaming Lv, Qilong Wang, and Peihua Li.
\newblock Joint distribution matters: Deep brownian distance covariance for
  few-shot classification.
\newblock In {\em IEEE/CVF Conference on Computer Vision and Pattern
  Recognition (CVPR)}, 2022.

\bibitem{xu2021learning}
Chengming Xu, Yanwei Fu, Chen Liu, Chengjie Wang, Jilin Li, Feiyue Huang, Li
  Zhang, and Xiangyang Xue.
\newblock Learning dynamic alignment via meta-filter for few-shot learning.
\newblock In {\em Proceedings of the IEEE/CVF Conference on Computer Vision and
  Pattern Recognition}, pages 5182--5191, 2021.

\bibitem{ye2020fewshot}
Han-Jia Ye, Hexiang Hu, De-Chuan Zhan, and Fei Sha.
\newblock Few-shot learning via embedding adaptation with set-to-set functions.
\newblock In {\em IEEE/CVF Conference on Computer Vision and Pattern
  Recognition (CVPR)}, pages 8808--8817, 2020.

\bibitem{Zhang_2020_CVPR}
Chi Zhang, Yujun Cai, Guosheng Lin, and Chunhua Shen.
\newblock Deepemd: Few-shot image classification with differentiable earth
  mover's distance and structured classifiers.
\newblock In {\em IEEE/CVF Conference on Computer Vision and Pattern
  Recognition (CVPR)}, 2020.

\bibitem{zhang2021meta}
Chi Zhang, Henghui Ding, Guosheng Lin, Ruibo Li, Changhu Wang, and Chunhua
  Shen.
\newblock Meta navigator: Search for a good adaptation policy for few-shot
  learning.
\newblock In {\em Proceedings of the IEEE/CVF International Conference on
  Computer Vision}, pages 9435--9444, 2021.

\bibitem{zhang2020iept}
Manli Zhang, Jianhong Zhang, Zhiwu Lu, Tao Xiang, Mingyu Ding, and Songfang
  Huang.
\newblock Iept: Instance-level and episode-level pretext tasks for few-shot
  learning.
\newblock In {\em International Conference on Learning Representations}, 2020.

\bibitem{zhang2021shallow}
Xueting Zhang, Debin Meng, Henry Gouk, and Timothy~M Hospedales.
\newblock Shallow bayesian meta learning for real-world few-shot recognition.
\newblock In {\em Proceedings of the IEEE/CVF International Conference on
  Computer Vision}, pages 651--660, 2021.

\bibitem{zhao2021looking}
Jiabao Zhao, Yifan Yang, Xin Lin, Jing Yang, and Liang He.
\newblock Looking wider for better adaptive representation in few-shot
  learning.
\newblock In {\em Proceedings of the AAAI Conference on Artificial
  Intelligence}, volume~35, pages 10981--10989, 2021.

\bibitem{zhou2021_ibot}
Jinghao Zhou, Chen Wei, Huiyu Wang, Wei Shen, Cihang Xie, Alan Yuille, and Tao
  Kong.
\newblock ibot: Image bert pre-training with online tokenizer.
\newblock {\em International Conference on Learning Representations (ICLR)},
  2022.

\bibitem{zhou2021binocular}
Ziqi Zhou, Xi Qiu, Jiangtao Xie, Jianan Wu, and Chi Zhang.
\newblock Binocular mutual learning for improving few-shot classification.
\newblock In {\em Proceedings of the IEEE/CVF International Conference on
  Computer Vision}, pages 8402--8411, 2021.

\bibitem{zintgraf2019fast}
Luisa Zintgraf, Kyriacos Shiarli, Vitaly Kurin, Katja Hofmann, and Shimon
  Whiteson.
\newblock Fast context adaptation via meta-learning.
\newblock In {\em International Conference on Machine Learning}, pages
  7693--7702. PMLR, 2019.

\end{thebibliography}


\begin{thebibliography}{10}\itemsep=-1pt

\bibitem{bertinetto2019_cifarfsl}
Luca Bertinetto, Joao~F Henriques, Philip Torr, and Andrea Vedaldi.
\newblock Meta-learning with differentiable closed-form solvers.
\newblock In {\em International Conference on Learning Representations}, 2019.

\bibitem{chen2019closerfewshot}
Wei-Yu Chen, Yen-Cheng Liu, Zsolt Kira, Yu-Chiang Wang, and Jia-Bin Huang.
\newblock A closer look at few-shot classification.
\newblock In {\em International Conference on Learning Representations}, 2019.

\bibitem{chen2021_mocov3}
Xinlei Chen, Saining Xie, and Kaiming He.
\newblock An empirical study of training self-supervised vision transformers.
\newblock In {\em Proceedings of the IEEE/CVF International Conference on
  Computer Vision}, pages 9640--9649, 2021.

\bibitem{chen2021pareto}
Zhengyu Chen, Jixie Ge, Heshen Zhan, Siteng Huang, and Donglin Wang.
\newblock Pareto self-supervised training for few-shot learning.
\newblock In {\em Proceedings of the IEEE/CVF Conference on Computer Vision and
  Pattern Recognition}, pages 13663--13672, 2021.

\bibitem{doersch2020_crosstransformers}
Carl Doersch, Ankush Gupta, and Andrew Zisserman.
\newblock Crosstransformers: spatially-aware few-shot transfer.
\newblock {\em Advances in Neural Information Processing Systems},
  33:21981--21993, 2020.

\bibitem{dosovitskiy2020_vit}
Alexey Dosovitskiy, Lucas Beyer, Alexander Kolesnikov, Dirk Weissenborn,
  Xiaohua Zhai, Thomas Unterthiner, Mostafa Dehghani, Matthias Minderer, Georg
  Heigold, Sylvain Gelly, et~al.
\newblock An image is worth 16x16 words: Transformers for image recognition at
  scale.
\newblock {\em arXiv preprint arXiv:2010.11929}, 2020.

\bibitem{gidaris2019boosting}
Spyros Gidaris, Andrei Bursuc, Nikos Komodakis, Patrick P{\'e}rez, and Matthieu
  Cord.
\newblock Boosting few-shot visual learning with self-supervision.
\newblock In {\em Proceedings of the IEEE/CVF International Conference on
  Computer Vision}, pages 8059--8068, 2019.

\bibitem{krizhevsky2009_cifar100}
Alex Krizhevsky, Geoffrey Hinton, et~al.
\newblock Learning multiple layers of features from tiny images.
\newblock 2009.

\bibitem{liu2021_swin}
Ze Liu, Yutong Lin, Yue Cao, Han Hu, Yixuan Wei, Zheng Zhang, Stephen Lin, and
  Baining Guo.
\newblock Swin transformer: Hierarchical vision transformer using shifted
  windows.
\newblock In {\em Proceedings of the IEEE/CVF International Conference on
  Computer Vision}, pages 10012--10022, 2021.

\bibitem{luo2021rectifying}
Xu Luo, Longhui Wei, Liangjian Wen, Jinrong Yang, Lingxi Xie, Zenglin Xu, and
  Qi Tian.
\newblock Rectifying the shortcut learning of background for few-shot learning.
\newblock {\em Advances in Neural Information Processing Systems}, 34, 2021.

\bibitem{mangla2020charting}
Puneet Mangla, Nupur Kumari, Abhishek Sinha, Mayank Singh, Balaji
  Krishnamurthy, and Vineeth~N Balasubramanian.
\newblock Charting the right manifold: Manifold mixup for few-shot learning.
\newblock In {\em Proceedings of the IEEE/CVF Winter Conference on Applications
  of Computer Vision}, pages 2218--2227, 2020.

\bibitem{oreshkin2018_tadam_fc100}
Boris Oreshkin, Pau Rodr{\'\i}guez~L{\'o}pez, and Alexandre Lacoste.
\newblock Tadam: Task dependent adaptive metric for improved few-shot learning.
\newblock {\em Advances in Neural Information Processing Systems}, 31, 2018.

\bibitem{qi2021transductive}
Guodong Qi, Huimin Yu, Zhaohui Lu, and Shuzhao Li.
\newblock Transductive few-shot classification on the oblique manifold.
\newblock In {\em Proceedings of the IEEE/CVF International Conference on
  Computer Vision}, pages 8412--8422, 2021.

\bibitem{ravi2017_optfsl}
Sachin Ravi and Hugo Larochelle.
\newblock Optimization as a model for few-shot learning.
\newblock In {\em International Conference on Learning Representations}, 2017.

\bibitem{ren2018_tieredimagenet}
Mengye Ren, Eleni Triantafillou, Sachin Ravi, Jake Snell, Kevin Swersky,
  Joshua~B Tenenbaum, Hugo Larochelle, and Richard~S Zemel.
\newblock Meta-learning for semi-supervised few-shot classification.
\newblock {\em arXiv preprint arXiv:1803.00676}, 2018.

\bibitem{russakovsky_2015imagenet}
Olga Russakovsky, Jia Deng, Hao Su, Jonathan Krause, Sanjeev Satheesh, Sean Ma,
  Zhiheng Huang, Andrej Karpathy, Aditya Khosla, Michael Bernstein, et~al.
\newblock Imagenet large scale visual recognition challenge.
\newblock {\em International Journal of Computer Vision}, 115(3):211--252,
  2015.

\bibitem{rusu2018meta}
Andrei~A Rusu, Dushyant Rao, Jakub Sygnowski, Oriol Vinyals, Razvan Pascanu,
  Simon Osindero, and Raia Hadsell.
\newblock Meta-learning with latent embedding optimization.
\newblock In {\em International Conference on Learning Representations}, 2018.

\bibitem{snell2017prototypical}
Jake Snell, Kevin Swersky, and Richard Zemel.
\newblock Prototypical networks for few-shot learning.
\newblock {\em Advances in Neural Information Processing Systems}, 30, 2017.

\bibitem{touvron2021_deit}
Hugo Touvron, Matthieu Cord, Matthijs Douze, Francisco Massa, Alexandre
  Sablayrolles, and Herv{\'e} J{\'e}gou.
\newblock Training data-efficient image transformers \& distillation through
  attention.
\newblock In {\em International Conference on Machine Learning}, pages
  10347--10357. PMLR, 2021.

\bibitem{vinyals2016_miniimagenet}
Oriol Vinyals, Charles Blundell, Timothy Lillicrap, Daan Wierstra, et~al.
\newblock Matching networks for one shot learning.
\newblock {\em Advances in Neural Information Processing Systems},
  29:3630--3638, 2016.

\bibitem{DeepBDC-CVPR2022}
Jiangtao Xie, Fei Long, Jiaming Lv, Qilong Wang, and Peihua Li.
\newblock Joint distribution matters: Deep brownian distance covariance for
  few-shot classification.
\newblock In {\em IEEE/CVF Conference on Computer Vision and Pattern
  Recognition (CVPR)}, 2022.

\bibitem{ye2020fewshot}
Han-Jia Ye, Hexiang Hu, De-Chuan Zhan, and Fei Sha.
\newblock Few-shot learning via embedding adaptation with set-to-set functions.
\newblock In {\em IEEE/CVF Conference on Computer Vision and Pattern
  Recognition (CVPR)}, pages 8808--8817, 2020.

\bibitem{Zhang_2020_CVPR}
Chi Zhang, Yujun Cai, Guosheng Lin, and Chunhua Shen.
\newblock Deepemd: Few-shot image classification with differentiable earth
  mover's distance and structured classifiers.
\newblock In {\em IEEE/CVF Conference on Computer Vision and Pattern
  Recognition (CVPR)}, 2020.

\bibitem{zhang2021shallow}
Xueting Zhang, Debin Meng, Henry Gouk, and Timothy~M Hospedales.
\newblock Shallow bayesian meta learning for real-world few-shot recognition.
\newblock In {\em Proceedings of the IEEE/CVF International Conference on
  Computer Vision}, pages 651--660, 2021.

\bibitem{zhou2021_ibot}
Jinghao Zhou, Chen Wei, Huiyu Wang, Wei Shen, Cihang Xie, Alan Yuille, and Tao
  Kong.
\newblock ibot: Image bert pre-training with online tokenizer.
\newblock {\em International Conference on Learning Representations (ICLR)},
  2022.

\end{thebibliography}

}
% \newpage
% \appendix
% \setcounter{table}{0}
% \setcounter{figure}{0}
% % \renewcommand{\thetable}{\Alph{section}\arabic{table}}
% \renewcommand{\thetable}{A\arabic{table}}
% \renewcommand{\thefigure}{A\arabic{figure}}

% \input{08_Supplementary}

%%%%%%%%%%%%%%%%%%%%%%%%%%%%%%%%%%%%%%%%%%%%%%%%%%%%%%%%%%%%
% Required Checklist (for submission)

% \include{07_Checklist}

%%%%%%%%%%%%%%%%%%%%%%%%%%%%%%%%%%%%%%%%%%%%%%%%%%%%%%%%%%%%

%%%%%%%%%%%%%%%%%%%%%%%%%%%%%%%%%%%%%%%%%%%%%%%%%%%%%%%%%%%%

\end{document}

\setcitestyle{square}

\maketitle
\appendix
% \input{08_Supplementary}
%%%%%%%%%%%%%%%%%%%%%%%%%%%%%%%%%%%%%%%%%%%%%%%%%%%%%%%%%%%%

\section{Selecting helpful patches at inference time in 1-shot scenarios}
\begin{figure}[!h] 
    \begin{center}
        \includegraphics[width=1\linewidth]{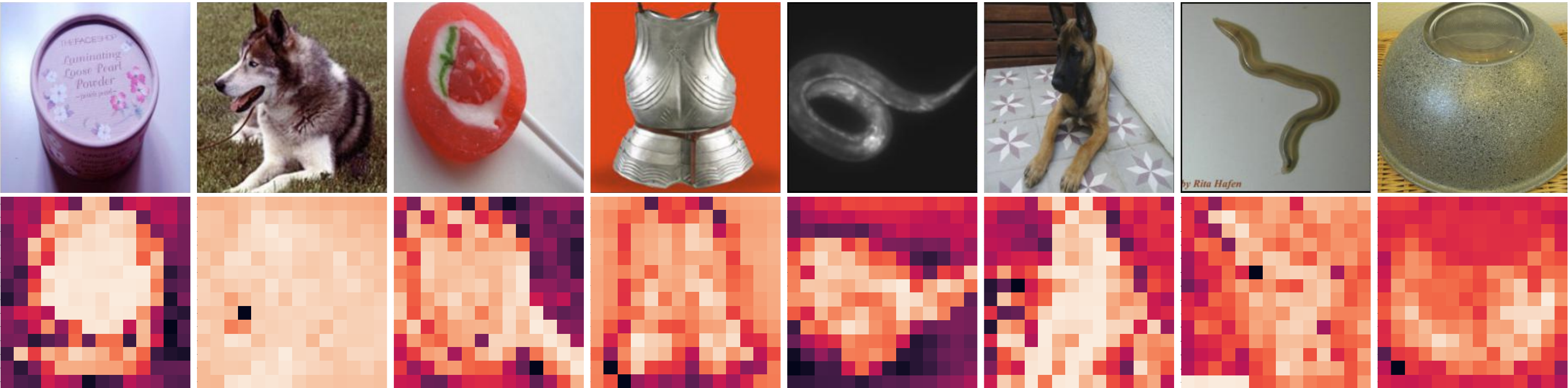}  
    \end{center}
    \caption{\textbf{Learning token importance at inference time.} Visualized importance weights learnt via online optimization for support set samples in a $5$-way $1$-shot task on the \textit{mini}ImageNet test set.} 
    \label{fig:sup_selfv_1shot}
\end{figure}

Figure~6 in the main paper demonstrates that our approach is able to successfully learn at inference time which image regions should be considered to classify the unknown query images in a 5-way 5-shot scenario. We additionally present the visualization of the token importance weights for the query images of a 5-way 1-shot scenario in \cref{fig:sup_selfv_1shot}. It can be clearly observed that the brighter regions representing higher importance of the respective image patches strongly relate to the actual objects that are to be classified, even in the case of smaller objects (2nd and 4th from the right). While our method only has access to significantly less information in the here presented 1-shot than in the case of 5-shot scenarios (see details in Section~$2.4$), our proposed way of masking the neighborhood of each pixel during the online optimization procedure still enables selection of the most helpful areas characteristic for the respective classes. 

\section{Discussion on model size and performance}
Related works have shown that model size seems to not be a good indicator for few-shot performance, most likely since training datasets are comparably small (\eg $38.4$K images in \textit{mini}ImageNet~\cite{vinyals2016_miniimagenet} vs. standard ImageNet with $1.28$M~\cite{russakovsky_2015imagenet}) and big networks are thus much more prone to overfit. Chen~\etal~\cite{chen2019closerfewshot} demonstrate in Figure~3 of their paper that the performance gains due to larger backbones plateau across all methods for backbones bigger than ResNet10 in their experiments and only offer diminishing gains (if any at all). The investigations of Mangla~\etal~\cite{mangla2020charting} yielded similar results, where the performance on the \textit{mini}ImageNet and \textit{tiered}ImageNet datasets even decreased by around $0.5$-$1\%$ when scaling up from ResNet18 to ResNet34 (Table 2). We thus conclude that increased number of parameters on its own does not lead to better few-shot performance, and the tendency of many recent works to choose the established ResNet12 ($12.4$M) over bigger backbones is highly likely a result of this. 

To gauge the influence of model size in FewTURE, we additionally investigate the use of the significantly smaller ViT-tiny architecture with only 5M parameters~\cite{touvron2021_deit}. Results in \cref{tab:sup_modelsize} show that our method achieves a competitive accuracy of $81.10\%$ on the \textit{mini}ImageNet test dataset with less than one seventh of the number of parameters of a WRN-28-10, but is (in contrast to many other methods like \eg \cite{ye2020fewshot}) able to leverage increased model sizes to further boost performance.
\begin{table}[!htb]
    \centering
    \caption{Investigating model size and performance. Average classification accuracy on the \emph{mini}-ImageNet test set, evaluated in a 5-way 5-shot scenario with a ViT-small backbone.}
    \vspace{0.4em}
    % \scalebox{0.89}
    {
    \renewcommand{\arraystretch}{1.2}
    \begin{tabular}{l l r c}
    \specialrule{.2em}{.1em}{.1em}
    \textbf{Method}
    &\textbf{Backbone}
    &\textbf{\#Params}
    &\textbf{Test Accuracy}
    \\
    \toprule \toprule
    ProtoNet~\cite{snell2017prototypical}
    &ResNet-12
    &$\approx12.4$M
    &$79.46{\scriptstyle \pm 0.48}$
    \\
    FEAT~\cite{ye2020fewshot}
    &ResNet-12
    &$\approx12.4$M
    &$82.05{\scriptstyle \pm 0.14}$
    \\
    DeepEMD~\cite{Zhang_2020_CVPR}
    &ResNet-12
    &$\approx12.4$M
    &$82.41{\scriptstyle \pm 0.56}$
    \\
    COSOC~\cite{luo2021rectifying}
    &ResNet-12
    &$\approx12.4$M
    &$85.16{\scriptstyle \pm 0.42}$
    \\
    Meta DeepBDC~\cite{DeepBDC-CVPR2022}
    &ResNet-12
    &$\approx12.4$M
    &$84.46{\scriptstyle \pm 0.28}$
    \\
    \hline 
    LEO~\cite{rusu2018meta}
    &WRN-28-10
    &$\approx36.5$M
    &$77.59{\scriptstyle \pm 0.12}$
    \\
    CC+rot~\cite{gidaris2019boosting}
    &WRN-28-10
    &$\approx36.5$M
    &$79.87{\scriptstyle \pm 0.33}$
    \\
    FEAT~\cite{ye2020fewshot}
    &WRN-28-10
    &$\approx36.5$M
    &$81.11{\scriptstyle \pm 0.14}$
    \\
    PSST~\cite{chen2021pareto}
    &WRN-28-10
    &$\approx36.5$M
    &$80.64{\scriptstyle \pm 0.32}$
    \\
    MetaQDA~\cite{zhang2021shallow}
    &WRN-28-10
    &$\approx36.5$M
    &$84.28{\scriptstyle \pm 0.69}$
    \\
    OM~\cite{qi2021transductive}
    &WRN-28-10
    &$\approx36.5$M
    &$85.29{\scriptstyle \pm 0.41}$
    \\
    \hline
    \rowcolor{Apricot!20!White} FewTURE (ours)
    &ViT-Tiny
    &$\approx\;\:5.0$M
    &$81.10{\scriptstyle \pm 0.61}$
    \\
    \rowcolor{Apricot!20!White} FewTURE (ours)
    &ViT-Small
    &$\approx22.0$M
    &$84.51{\scriptstyle \pm 0.53}$
    \\
    \rowcolor{Apricot!20!White} FewTURE (ours)
    &Swin-Tiny
    &$\approx29.0$M
    &$\bf{86.38{\scriptstyle \pm 0.49}}$
    \\
    \hline
    \end{tabular}
    }
    \label{tab:sup_modelsize}
\end{table}

\section{Discussion on self-supervised vs. supervised pretraining}
\textbf{Performance in few-shot learning.} We demonstrate in Figure~4 of the main paper that self-supervised pretraining with masked image modelling as pretext task provides a significant advantage over supervised pretraining for our approach -- a finding that differs from prior non-few-shot literature where self-supervised methods only moderately outperform their supervised counterparts~\cite{zhou2021_ibot} or even perform worse in some cases~\cite{chen2021_mocov3}. We provide our interpretation and insights regarding this in the following.

Few-shot classification is distinctively different from `conventional' classification (like investigated in~\cite{chen2021_mocov3}) in one important aspect: novel previously unseen classes are encountered at test time. As such, supervised learning induces a tendency of the representation space to overfit to the structure of the classes observed during training. In other words, the representation space is created and condensed to easily separate observed training classes, but at the expense of distorting other dimensions that might be crucial to correctly distinguish yet unseen classes. This is known in the few-shot literature as ‘\textit{supervision collapse}’~\cite{doersch2020_crosstransformers}. Since no class labels are provided during the self-supervised pretraining, we expect the method to create a more general/less distorted representation space that is significantly better suited to generalize to yet unseen classes and avoid collapse. These intuitions are supported by the results we have obtained (Fig 4.). We further observe that self-supervised training is helpful to prevent early overfitting when learning from small few-shot datasets (e.g. 38.4K \textit{mini}ImageNet~\cite{vinyals2016_miniimagenet} vs. 1.2M ImageNet1K~\cite{russakovsky_2015imagenet}).

\textbf{Training details of supervised pretraining.} For adequate comparison to related work in few-shot learning, we follow the widely adopted pretraining scheme used in FEAT~\cite{ye2020fewshot} and other works (e.g. DeepEMD~\cite{Zhang_2020_CVPR}) for our supervised pretraining. In detail, we train the network with a cross-entropy loss on the training set of the respective dataset to solve a standard classification task (e.g. for \textit{mini}ImageNet: 64 classes) -- \ie, using the exact same data we use for self-supervised pretraining. Like~\cite{ye2020fewshot} we use the representations of the penultimate layer (before the classifier) to evaluate the performance and quality of the embeddings. To judge suitability of the encoder for few-shot tasks, an N-way 1-shot task is commonly solved (e.g. N=16 for \textit{mini}ImageNet due to the 16 classes in the validation set) -- and we tried three different variants here:
\begin{itemize}
    \item[1. \& 2.] One sample per class is encoded to produce a class-embedding (`prototype'), and classification performance is evaluated using 15 queries per class (as used in recent related works). To retrieve one embedding per sample, we use the average over all patch tokens produced by the Transformer architecture. 
    For fairness regarding metrics, we evaluate both:
    \begin{enumerate}
        \item embedding distance (MSE) and
        \item embedding similarity (cosine) to perform classification.
    \end{enumerate}
    \item[3.] We additionally use our own patch-based classifier to evaluate the few-shot setting using all patch embeddings (as we later do during fine-tuning \& evaluation).
\end{itemize}
We perform validation over 200 such few-shot tasks after every epoch during training and pick the best-performing model regarding highest average validation accuracy. We encountered clear signs of overfitting during this type of supervised training, with the training accuracy consistently improving to convergence, but validation accuracy plateauing (or decreasing) rather early on ($\sim$350-500ep), independent of the variant we used to evaluate on the validation set.

\section{Ablation studies on components of FewTURE}
In this section, we provide further insights into our approach and the design choices we made.
\subsection{Ablation on inner loop token reweighting}
A more detailed version of the average classification test accuracies achieved with a meta fine-tuned ViT backbone on the \textit{mini}ImageNet dataset used for the visualization of the contribution for different numbers of token reweighting steps during online optimisation (main paper, Figure~7) is presented in \cref{tab:update_steps_ablation_time_sup}, including the respective $95\%$ confidence intervals. 
As discussed in the main paper, we observed a strong initial increase of~$1.15\%$ when using our proposed adaptation via online optimization (steps$>0$). While a higher number of inner-loop updates seems to still lead to increased accuracy across all our test runs, this benefit brings along higher computational cost as can be seen in the second row of \cref{tab:update_steps_ablation_time_sup}. We generally found settings between 5 and 15 steps to be a good accuracy \textit{vs.} inference-time trade-off. Our experiments were conducted using an Nvidia-2080ti GPU and the stated inferences times have been averaged over 1800 query sample classifications. It is to be noted that the code has not been specifically optimized for fast inference times, and these values should rather be interpreted in a relative manner.

\begin{table}[!htb]
    \centering
    \caption{Average classification accuracy and inference times on the \emph{mini}-ImageNet test set for varying inner loop optimization steps, evaluated in a 5-way 5-shot scenario with a ViT-small backbone and SDG with 0.1 as learning rate. Experiments were conducted using an Nvidia-2080ti and runtimes were averaged over 1800 query sample classifications.}
    \vspace{0.4em}
    \scalebox{0.89}
    {
    \renewcommand{\arraystretch}{1.2}
    \begin{tabular}{l c c c c c}
    \specialrule{.2em}{.1em}{.1em}
    &\textbf{0 steps}
    &\textbf{5 steps}
    &\textbf{10 steps}
    &\textbf{15 steps}
    &\textbf{20 steps}
    \\
    \toprule \toprule
    \textbf{Accuracy}
    &$82.68{\scriptstyle \pm 0.59}$
    &$83.83{\scriptstyle \pm 0.59}$
    &$83.89{\scriptstyle \pm 0.57}$
    &$84.05{\scriptstyle \pm 0.55}$
    &$84.51{\scriptstyle \pm 0.53}$
    \\
    \textbf{Inference time} [ms]
    &$156.86{\scriptstyle \pm 2.16}$   % 156.23 +- 2.24  % 157.40 +- 2.07 % 156.95 +- 2.17
    &$159.86{\scriptstyle \pm 2.12}$
    &$162.11{\scriptstyle \pm 2.11}$   % 162.11 +- 1.80  % 162.02 +- 2.25  % 162.21 +- 2.30
    &$165.62{\scriptstyle \pm 2.06}$   % 164.70 +- 2.15  % 165.14 +- 2.18  % 167.01 +- 1.85
    &$168.62{\scriptstyle \pm 2.22}$   % 167.60 +- 2.18  % 169.39 +- 2.15  % 168.87 +- 2.32
    \\
    \hline
    \end{tabular}
    }
    \label{tab:update_steps_ablation_time_sup}
\end{table}

\subsection{Ablation on token aggregation and similarity metrics}
As discussed in the main paper, we use the \textit{logsumexp} operation to aggregate our similarity logits as it poses a rigorous and numerically stable way of combining individual class probabilities (one for each token) to a valid overall probability distribution over classes for each image, independent of how the individual token (log) probability scores are obtained. Table~\ref{tab:sup_abl_aggsim}\,(\subref{subtab:sup_abl_aggregation}) shows the results of additional experiments (training and testing) using our method (ViT-small) and 15 token reweighting steps with the only change being aggregation of the logtis via \textit{mean}, and we found it to underperform our chosen \textit{logsumexp} method of aggregation. Direct addition without normalization (\ie just summing up all logits) proved unstable due to large logit values and was thus not included in this table.

We further investigated the use of alternate metrics to compute the similarity between different tokens. Both the use of the negative Euclidean distance and unscaled dot-product yielded inferior results compared to the temperature-scaled cosine distance we use in FewTURE (Table~\ref{tab:sup_abl_aggsim}\,(\subref{subtab:sup_abl_simmetrics})).

\begin{table}[!htb]
    \caption{Ablation on token aggregation method and similarity metric. Reported are the average classification accuracies on the \emph{mini}ImageNet test set evaluated in a 5-way 5-shot scenario with a ViT-small backbone.}\label{tab:sup_abl_aggsim}
    \begin{subtable}{.5\linewidth}
      \centering
        \caption{Token aggregation}\label{subtab:sup_abl_aggregation}
        \begin{tabular}{lc}
            \specialrule{.2em}{.1em}{.1em}
            \textbf{Aggregation method} & \textbf{Test Accuracy}\\
            \toprule\toprule
            logsumexp & $\mathbf{84.05\pm0.53}$ \\ 
            mean logits & $80.13\pm0.60$ \\  
            \bottomrule
        \end{tabular}
    \end{subtable}%
    \begin{subtable}{.5\linewidth}
      \centering
        \caption{Similarity metrics}\label{subtab:sup_abl_simmetrics}
        \begin{tabular}{lc}
        \specialrule{.2em}{.1em}{.1em}
            \textbf{Metric} & \textbf{Test Accuracy}\\
            \toprule\toprule
            cosine similarity & $\mathbf{84.05\pm0.53}$ \\ 
            neg. Euclidean dist. & $81.85\pm0.58$ \\  
            unscaled dot-prod. & $37.60\pm0.64$ \\  
        \bottomrule
        \end{tabular}
    \end{subtable} 
\end{table}

\subsection{Ablation regarding temperature scaling of embedding similarity logits}
\label{sec:sup_temp}
As reported in the main paper, we use the temperature~$\tau_S$ to rescale the logits of our task-specific similarity matrix~$\Tilde{\boldsymbol{S}}$ via division (or the original similarity matrix~$\boldsymbol{S}$ in case no task-specific adaptation shall be used). We investigate two different ways of temperature scaling: (i) the possibility of using a fixed temperature defined as $\nicefrac{1}{\sqrt{d}}$ where $d$ is the dimension of the patch embeddings of the respective architecture, and (ii) learning the appropriate temperature during the meta fine-tuning procedure. In practice, we learn~$\log(\tau_S)$ to ensure $\tau_S\geq0$.

\begin{figure}[h]
\centering
    \begin{subfigure}[b]{0.48\textwidth}     
        \centering
         \includegraphics[width=\textwidth]{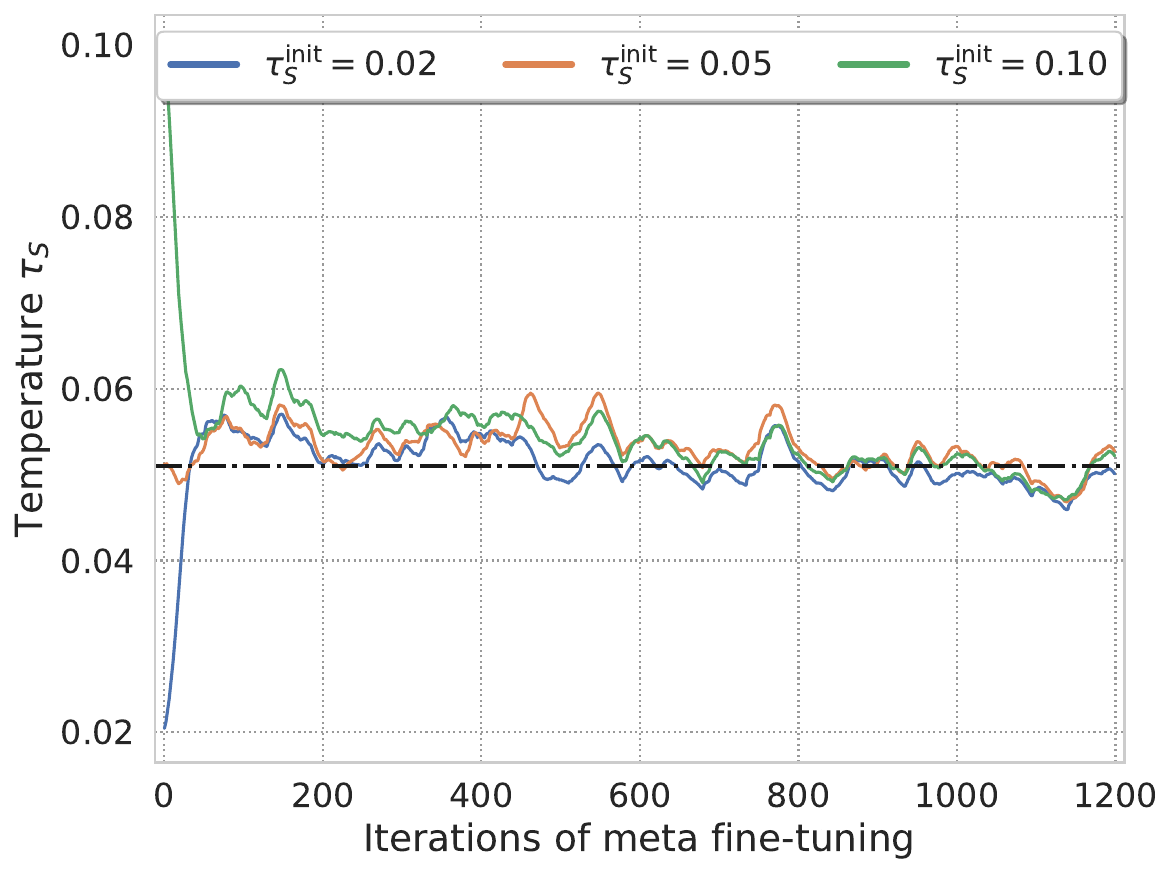}
         \caption{}
         \label{subfig:temp_ablation_sup_mini1}
     \end{subfigure}
     \hfill
     \begin{subfigure}[b]{0.48\textwidth}
         \centering
         \includegraphics[width=\textwidth]{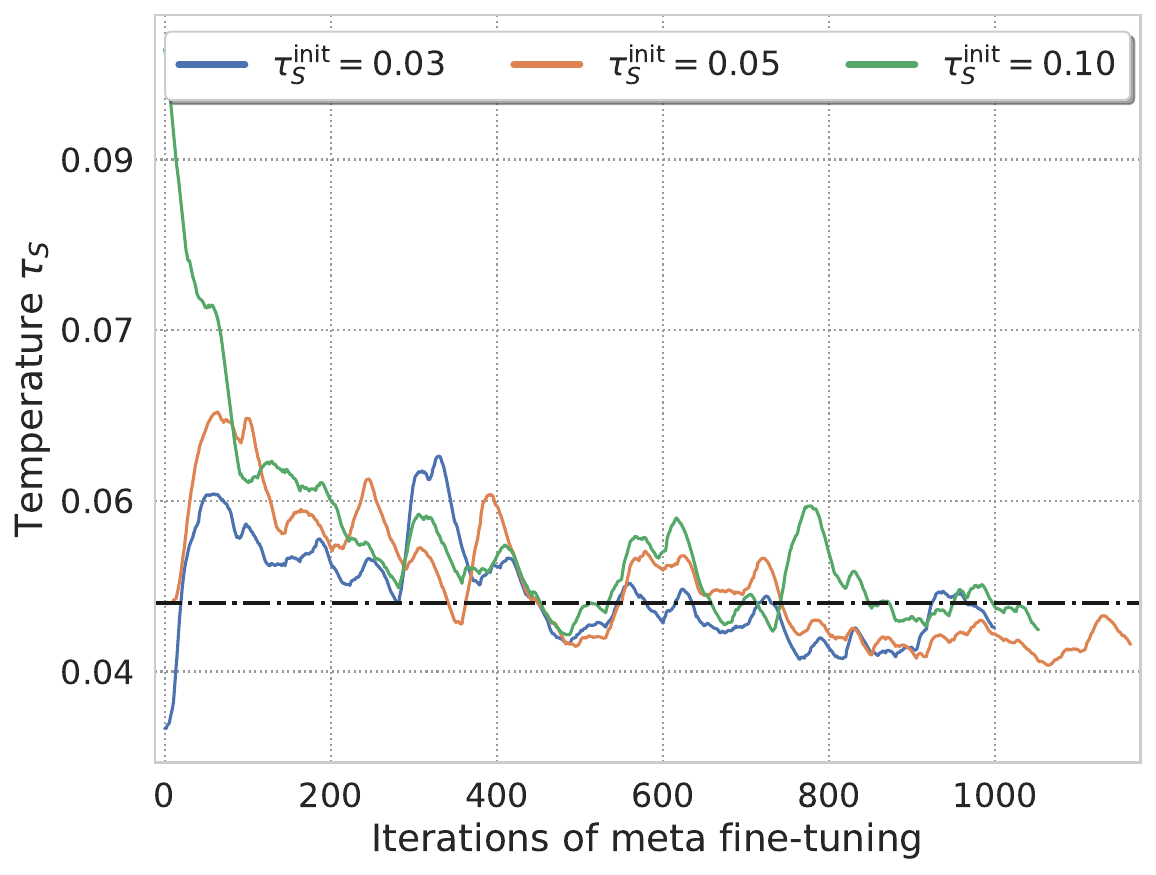}
         \caption{}
         \label{subfig:temp_ablation_sup_tiered1}
     \end{subfigure}
     \begin{subfigure}[b]{0.48\textwidth}     
        \centering
         \includegraphics[width=\textwidth]{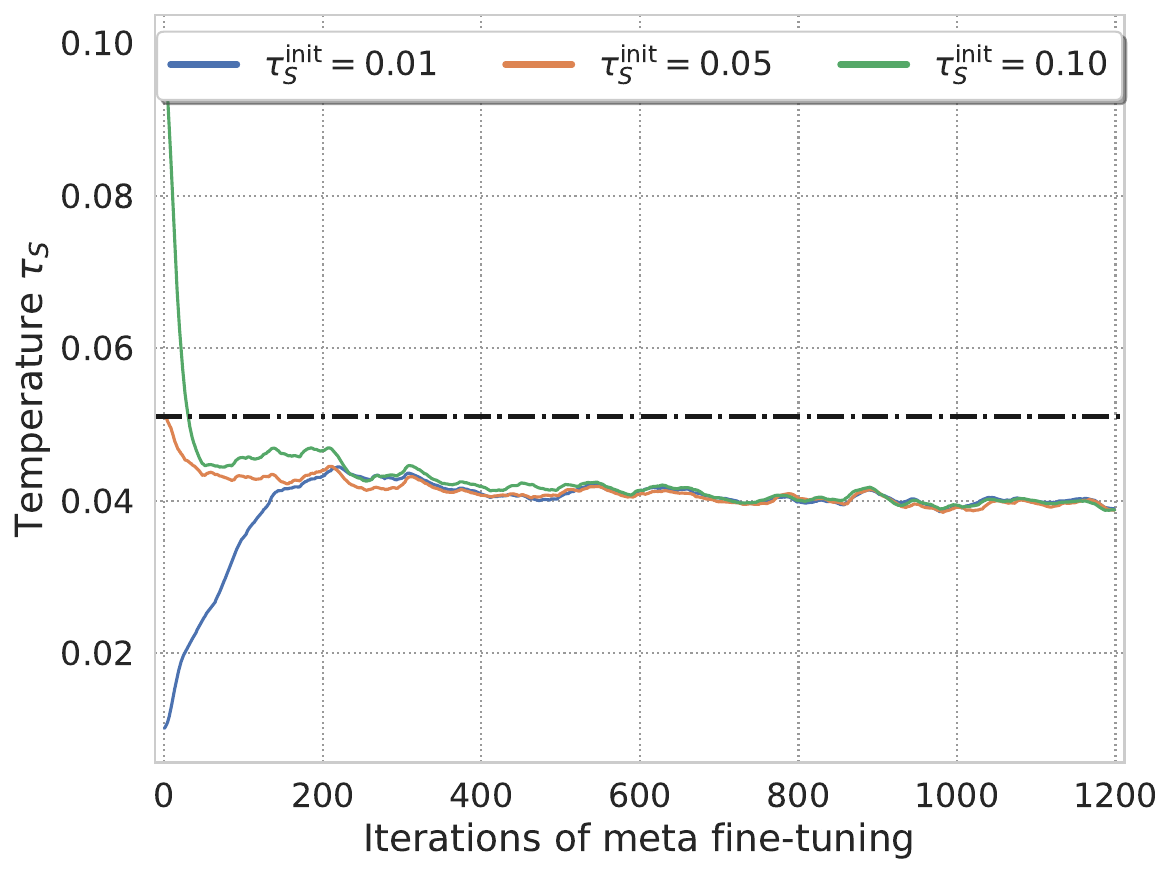}
         \caption{}
         \label{subfig:temp_ablation_sup_mini5}
     \end{subfigure}
     \hfill
     \begin{subfigure}[b]{0.48\textwidth}
         \centering
         \includegraphics[width=\textwidth]{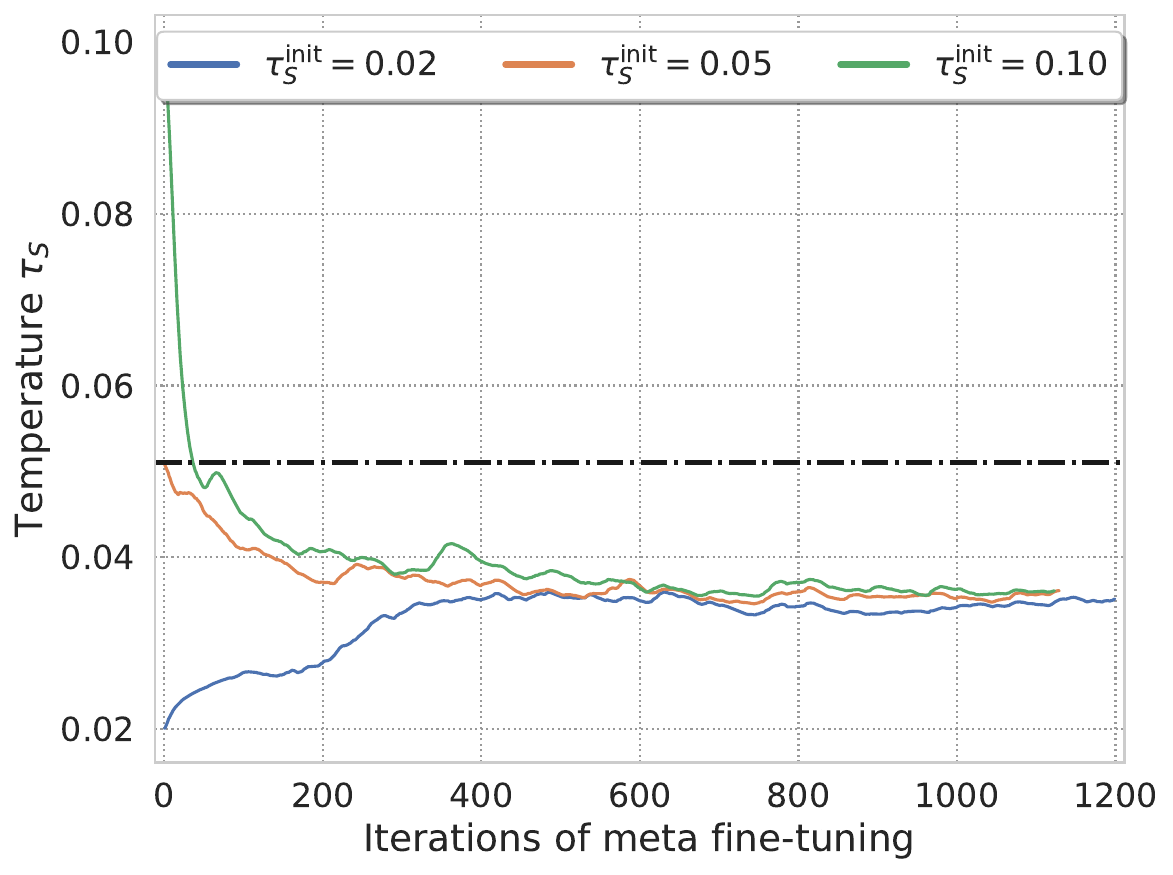}
         \caption{}
         \label{subfig:temp_ablation_sup_tiered5}
     \end{subfigure}
     \caption{\textbf{Temperature for rescaling similarity logits.} (\subref{subfig:temp_ablation_sup_mini1}) and (\subref{subfig:temp_ablation_sup_tiered1}) show the learned temperatures for 5-way 1-shot scenarios on \textit{mini}ImageNet and \textit{tiered}ImageNet, respectively. The corresponding 5-way 5-shot results are depicted in (\subref{subfig:temp_ablation_sup_mini5}) and (\subref{subfig:temp_ablation_sup_tiered5}). All experiments have been conducted using a ViT-small architecture.}
     \label{fig:temp_ablation_sup}
\end{figure}

We observe throughout our 1-shot experiments depicted in \cref{fig:temp_ablation_sup}\,(\subref{subfig:temp_ablation_sup_mini1}) and (\subref{subfig:temp_ablation_sup_tiered1}) that the temperature converges towards our default values of $\nicefrac{1}{\sqrt{d}}$ shown as a dashed horizontal line. This is independent of the initial value of the temperature parameter~$\tau_S^{\mathrm{init}}$. For the 5-way 5-shot experiments presented in \cref{fig:temp_ablation_sup}\,(\subref{subfig:temp_ablation_sup_mini5}) and (\subref{subfig:temp_ablation_sup_tiered5}) however, we observe that while our default value still achieves good results, the learned temperature converges to a slightly lower value across all experiments. % -- which could indicate the slightly higher confidence and consideration of a smaller local neighborhood for such scenarios. 

\subsection{Development over the course of pretraining}
We further present insights into the development of the accuracy during self-supervised pretraining. Since our pretraining procedure is entirely unsupervised and does hence not include any labels, we investigate models trained for a variety of different epochs and evaluate these on the test set using the proposed similarity-based classification method \textit{with} (`5 steps' and `15 steps') and \textit{without} (`None') and present the results in \cref{tab:sup_temp}. Note that no meta fine-tuning was employed here. We observe that while the performance significantly increases over the first 50 epochs, there seems to be some saturation and even slight decrease in performance until above 500 epochs where the accuracy increases again and (mostly) achieves highest results in this study. 

\begin{table}[!htb]
    \centering
    \caption{\textbf{Development of test accuracy in self-supervised pretraining.} Results obtained for a 5-way 5-shot scenario on the \textit{tiered}ImageNet test set using our proposed classifier with a ViT-small backbone. For online optimisation (\ie, steps$>0$), we use SGD with 0.1 as learning rate.}
    \vspace{0.4em}
    \scalebox{0.89}
    {
    \renewcommand{\arraystretch}{1.2}
    \begin{tabular}{@{}l *7c@{}}
    \specialrule{.2em}{.1em}{.1em}
    & \textbf{Reweighting}
    & \multicolumn{6}{c}{\textbf{Epochs}}
    \\
    &\textbf{steps} 
    &\textbf{1}
    &\textbf{50}
    &\textbf{100}
    &\textbf{250}
    &\textbf{500}
    &\textbf{800}
    \\
    \toprule \bottomrule 
    % \multirow{3}{*}{\STAB{\rotatebox[origin=c]{90}{ViT}}} 
    & None 
    &$39.20{\scriptstyle \pm 0.69}$
    &$73.30{\scriptstyle \pm 0.75}$ 
    &$73.63{\scriptstyle \pm 0.73}$  
    &$72.84{\scriptstyle \pm 0.72}$ 
    &$71.51{\scriptstyle \pm 0.72}$ 
    &$73.83{\scriptstyle \pm 0.74}$ 
    \\        % 0.15863254883334776 +- 0.001959822071136903 seconds.
    & 5 steps % 0.1584767076043257 +- 0.002114180813224711 seconds.
    &$39.34{\scriptstyle \pm 0.69}$ % 0.15541064947667868 +- 0.001998311336936429 seconds
    &$73.59{\scriptstyle \pm 0.74}$ % 0.1606391878149907 +- 0.001973436025593111 seconds.
    &$74.03{\scriptstyle \pm 0.73}$ % 0.15961783906166147 +- 0.002261641829541441 seconds
    &$73.10{\scriptstyle \pm 0.73}$ 
    &$71.82{\scriptstyle \pm 0.72}$ % 0.15928935435498412 +- 0.002263404328243018 seconds
    &$74.16{\scriptstyle \pm 0.73}$ % 0.15986696157332517 +- 0.002122876708470918 seconds
    \\
    & 15 steps
    &$39.43{\scriptstyle \pm 0.69}$ 
    &$73.86{\scriptstyle \pm 0.73}$ 
    &$74.48{\scriptstyle \pm 0.74}$ 
    &$73.41{\scriptstyle \pm 0.75}$ 
    &$72.16{\scriptstyle \pm 0.73}$ 
    &$74.42{\scriptstyle \pm 0.74}$ 
    \\
    % \midrule
    % \multirow{2}{*}{\STAB{\rotatebox[origin=c]{90}{Swin}}} 
    % & None 
    % &$-.-{\scriptstyle \pm 0.-}$ 
    % &$78.40{\scriptstyle \pm 0.69}$  % 
    % &$80.90{\scriptstyle \pm 0.66}$  % 
    % &$83.96{\scriptstyle \pm 0.62}$  % 
    % &$68.05{\scriptstyle \pm 0.97}$  % 
    % &$82.81{\scriptstyle \pm 0.64}$  % 
    % \\
    % & 15 steps
    % &$40.58{\scriptstyle \pm 0.68}$  %
    % &$79.47{\scriptstyle \pm 0.69}$  %
    % &$81.89{\scriptstyle \pm 0.65}$  %
    % &$84.77{\scriptstyle \pm 0.61}$  %
    % &$84.03{\scriptstyle \pm 0.64}$  %
    % &$83.58{\scriptstyle \pm 0.64}$  %
    % \\
    \bottomrule
    \end{tabular}
    }
    \label{tab:sup_temp}
\end{table}

\section{Further visualization of instance embeddings}
\begin{figure}[t] 
    \begin{center}
        \includegraphics[width=1\linewidth]{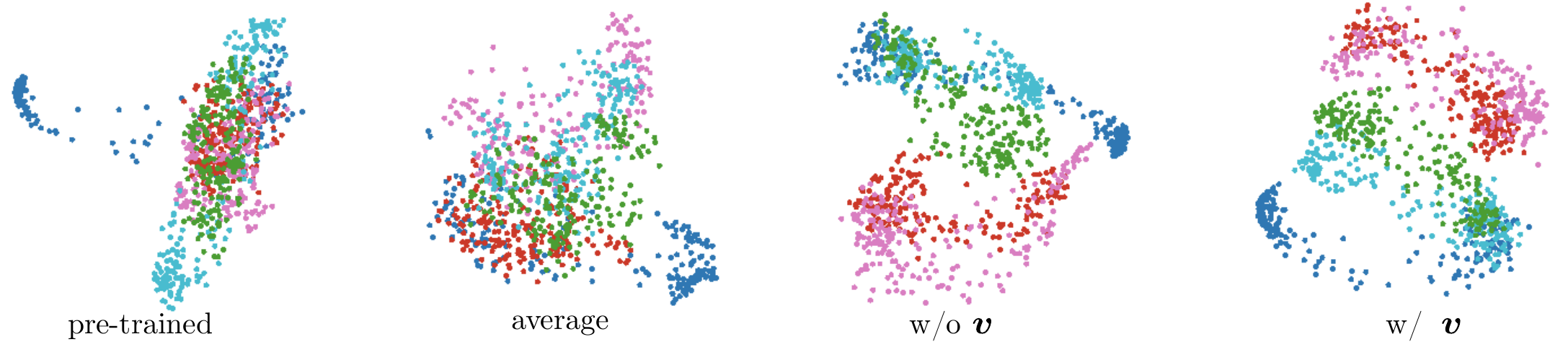}  
    \end{center}
    \caption{\textbf{Instance embeddings after meta fine-tuning.} Visualized are the projected tokens of 5 instances of the same novel support set class for different meta fine-tuning (M-FT) methods (after self-supervised pretraining). From left to right: self-supervised pretraining only, M-FT using an average embedding per class, M-FT using our classifier but without task-specific token reweighting, M-FT using our classifier with 15 reweighting steps. (Projection via PCA to main dimensions.)}
    \label{fig:TSNE_sup}
\end{figure}

Figure~5 in the main paper depicts instance and class embeddings visualized via PCA projection to the three dominant dimensions. Figure~\ref{fig:TSNE_sup} additionally depicts a comparison of projected views of the tokens of 5 instances from a novel class in embedding space for different ways of meta training. While the representations obtained from the network meta fine-tuned by using common averaging over the embeddings (`\textit{average}') do not exhibit any clear separation of the instances, the embeddings obtained with our classifier seem to retain the instance information (`\textit{w/o} $\boldsymbol{v}$') and separation is improved when using token importance reweighting (`\textit{w/} $\boldsymbol{v}$'). These results indicate that our similarity-based classifier coupled with task-specific token reweighting is able to better disentangle the embeddings of different instances from the same class, which further prevents the network from supervision collapse and helps to achieve the higher performance observed on the benchmarks.

\section{Datasets used for evaluation}
We train and evaluate our approach presented in the main paper on the following few-shot image classification datasets: \par
\textbf{\textit{mini}ImageNet}. The \textit{mini}ImageNet dataset has been initially proposed by~\cite{vinyals2016_miniimagenet} with follow-up modifications by~\cite{ravi2017_optfsl} and consists of a specific 100 class subset of ImageNet~\cite{russakovsky_2015imagenet} with 600 images for each class. The data is split into 64 training, 16 validation and 20 test classes. \par
\textbf{\textit{tiered}ImageNet}. Similar to the previous dataset, the \textit{tiered}ImageNet~\cite{ren2018_tieredimagenet} is a subset of classes selected form the bigger ImageNet~\cite{russakovsky_2015imagenet} dataset, however with a substantially larger set of classes and different structure in mind. It comprises a selection of 34 super-classes with a total of 608 categories, totalling in $779{,}165$ images that are split into 20,6 and 8 super-classes to achieve better separation between training, validation and testing, respectively. \par
\textbf{CIFAR-FS}. The CIFAR-FS dataset~\cite{bertinetto2019_cifarfsl} contains the 100 categories with 600 images per category from the CIFAR100~\cite{krizhevsky2009_cifar100} dataset which are split into 64 training, 16 validation and 20 test classes.\par
\textbf{FC-100}. The FC-100 dataset~\cite{oreshkin2018_tadam_fc100} is also derived from CIFAR100~\cite{krizhevsky2009_cifar100} but follows a splitting strategy similar to \textit{tiered}ImageNet to increase difficulty through higher separation, resulting in 60 training, 20 validation and 20 test classes.

\section{Implementation details}
We present further details regarding our implementation and used hyperparameters in the following.
\subsection{Pretraining}
\textbf{GPU usage}.
We pretrain our models with the use of 4 Nvidia A100 GPUs with 40GB each for our ViT~\cite{dosovitskiy2020_vit,touvron2021_deit} and 8 such GPUs for our Swin~\cite{liu2021_swin} variants. \par
\textbf{Hyperparameter choice}. 
We follow the strategy introduced by~\cite{zhou2021_ibot} to pretrain our Transformer backbones and mostly stick to the hyperparameter settings reported in their work. We generally use two global crops and 10 local crops with crop scales of $(0.4, 1.0)$ and $(0.05, 0.4)$, respectively. We further use a patch size of 16 for our ViT models and a window size of 7 for Swin, corresponding to the default sizes for ViT-small~\cite{dosovitskiy2020_vit,touvron2021_deit} and Swin-tiny~\cite{liu2021_swin}. We use an output dimension of $8192$ for the projection heads across all models, and employ random Masked Image Modelling with prediction ratios $(0, 0.3)$ and variances $(0, 0.2)$. Our ViT and Swin architectures are trained with an image size of $224\times 224$ arranged in batches of size $512$ samples for $1600$ and $800$ epochs, respectively, using a linearly ramped-up learning rate (over first 10 epochs) of $5\mathrm{e-}4 \times \mathrm{batch size}/256$. 
For detailed information, we would like to refer the interested reader to the work by Zhou \etal~\cite{zhou2021_ibot} where more background information regarding the influence and justification of these hyperparameters is provided.
\subsection{Meta fine-tuning}
\textbf{GPU usage}. 
During the meta fine-tuning (M-FT) stage, we use 1 and 2 Nvidia 2080-ti GPUs for ViT-small and Swin-tiny, respectively, across all 4 datasets. 

\textbf{Hyperparameters}. 
We fix the input image size as $224\times224$ for all datasets.
We use the SGD optimizer along with a learning rate of $2\mathrm{e-}4$, $0.9$ as the momentum value and $5\mathrm{e-}4$ as the weight decay. Additionally, we employ a learning rate scheduler with cosine annealing for $5{,}000$ iterations as one cycle, ramping down to $5\mathrm{e-}5$ at the end of each cycle. 

\textbf{Online optimization}.
During the online learning of the token importance reweighting vectors, we adopt the SGD optimizer with 0.1 as the learning rate. For online update steps, we generally choose a default value of 15 steps across all datasets. For further details regarding the temperature scaling procedure used to rescale our task-specific similarity logits, please refer to Section~\ref{sec:sup_temp}.

{
\small
\bibliographystyle{ieee_fullname}
\bibliography{patchfs}
}

%%%%%%%%%%%%%%%%%%%%%%%%%%%%%%%%%%%%%%%%%%%%%%%%%%%%%%%%%%%%